%% file: main.tex
\documentclass[10pt,twocolumn,letterpaper]{article}
\usepackage[pagenumbers]{cvpr}
\input{preamble}
\usepackage{array}
\definecolor{cvprblue}{rgb}{0.21,0.49,0.74}
\usepackage[pagebackref,breaklinks,colorlinks,allcolors=cvprblue]{hyperref}
\def\paperID{*****}
\def\confName{CVPR}
\def\confYear{2027}

\title{TileShare: Training-Free Acceleration of Overlapping Tiled ViT Decoding}
\author{Xiangzhong Ye \qquad Zhongshen Li \qquad Shen You \qquad Ziyang Meng \qquad Liang Xu\\ Junkai Ji \qquad Qiuzhen Lin \qquad Xiangtao Li \qquad Ka-Chun Wong$^{\dagger}$}

\begin{document}
\twocolumn[{%
\renewcommand\twocolumn[1][]{#1}%
\maketitle
\input{fig/teaser}}]
{\renewcommand{\thefootnote}{}\footnotetext{$^{\dagger}$Corresponding author.}}
\input{sec/0_abstract}
\input{sec/1_intro}
\input{sec/2_related}
\input{sec/3_method}
\input{sec/4_mechanism}
\input{sec/5_experiments}
\input{sec/6_learn}
\input{sec/7_conclusion}
{
    \small
    \bibliographystyle{ieeenat_fullname}
    \bibliography{refs}
}

\appendix
\renewcommand{\thetable}{S\arabic{table}}
\renewcommand{\thefigure}{S\arabic{figure}}
\setcounter{table}{0}
\setcounter{figure}{0}
\twocolumn[{\centering\Large\bfseries Supplementary Material\par\vspace{1.5em}}]
\input{sec/supp_body}
\end{document}

%% file: preamble.tex
\usepackage{cuted}

\usepackage{currfile}

\usepackage{caption}

\usepackage{xspace}
\usepackage{booktabs}
\usepackage{multirow}
\usepackage{amsmath,amssymb}
\usepackage{graphicx}
\newcommand{\method}{TileShare\xspace}

\newcommand{\splitk}[1]{\mbox{TileShare-#1}\xspace}
\newcommand{\ttt}[1]{\texttt{#1}}
\usepackage{iftex}
\ifXeTeX
  \usepackage{fontspec}
\fi

%% file: fig/teaser.tex
\begin{center}
\centering
\captionsetup{type=figure}
\includegraphics[width=\textwidth]{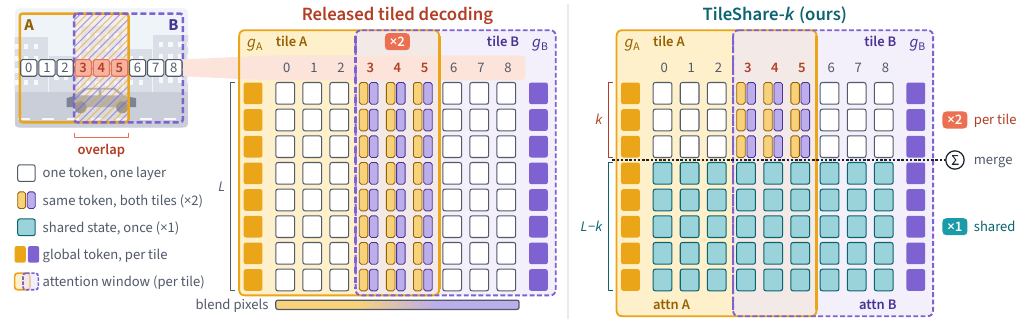}
\vspace{-1.0em}
\captionof{figure}{\textbf{Overlapping tiles as token $\times$ layer grids} (toy: two tiles, $k{=}3$). Each cell is one token's per-token operations in one layer (Norm, QKV, out-projection, MLP); shaded frames are the tiles' attention windows, each with its own global token $g$. \textbf{Left:} the released decoder runs both tiles in full; overlap tokens are thus computed twice in every layer. \textbf{Right:} \method keeps the attention windows (still overlapping) but computes every other operation once per token; only the first $k$ layers stay per tile.}
\label{fig:method}\label{fig:teaser}
\end{center}
\vspace{0.3em}

%% file: sec/0_abstract.tex
\begin{abstract}
Fixed-window Vision Transformer (ViT) decoders run at deployment resolution as overlapping tiles: every token in an overlap band is processed by several full decoder instances. In the 2.4B-parameter MiniMax-H3 video decoder, about half of all per-token computation is duplicated.
Removing the tiles fails because attention, global tokens and coordinates are bound to them.
\method keeps each released tile's attention and computes every other operation once per token. The released computation changes at one kind of step only: an overlap token's per-tile results are blended into one shared state with the released weights.
Interventions show that this blend costs quality where attention reads the shared states, mainly in the first few layers, which \method therefore keeps per tile.
Without training, keeping six layers per tile decodes $1.38$--$1.51\times$ faster than batched and compiled tiling on two GPU types (up to $1.96\times$ with FP8) and is equivalent to the released decoder in PSNR, LPIPS and DISTS on DAVIS videos. On unseen UVG videos with the setting frozen beforehand, it matches PSNR and DISTS and improves LPIPS.
\method carries over to two further decoders and is again confirmed on unseen data. It is equivalent in PSNR and DISTS on MAGI-1. On the ViTok-v2 image autoencoder at 2K--4K, it matches overlapping tiling in PSNR, improves LPIPS and DISTS and decodes $1.5$--$1.8\times$ faster.
How many layers stay per tile is measured once per model (for decoders, without labels); we report it for 14 decoders and encoders.
Optionally, two GPU-hours of self-distillation let MiniMax-H3 share every layer without a loss in PSNR, and a coordinate field trained while the network stays frozen brings fully shared CLIP-based open-vocabulary segmentation to parity with sliding windows on seven benchmarks.
\end{abstract}

%% file: sec/1_intro.tex
\section{Introduction}
\label{sec:intro}

Latent video and image generators map latent representations back to pixels with an autoencoder decoder; pure Vision Transformer (ViT) decoders are now prevalent~\cite{minimaxh3,magi1,vitok,atoken}. They are trained on fixed windows. Therefore, at deployment resolution, they run as \emph{overlapping tiled inference}: each overlapping tile is decoded on its own and the outputs are blended with linear ramps. As generators are distilled to a few denoising steps, decoding becomes a visible share of latency: 16\% of a 4-step text-to-video request at 768p, and 40\% when a single denoiser pass suffices.

Overlapping tiled inference is an ensemble: every token in an overlap band is processed by several independent decoder instances, whose outputs are then averaged (\cref{fig:teaser}, left). In the 2.4B-parameter MiniMax-H3 decoder~\cite{minimaxh3}, spatial overlap processes each token 2.03 times on average at 704$\times$1280; about half of all per-token computation is thus duplicated.

The obvious ways to remove this duplication lose quality for two different reasons (\cref{sec:tile}). Larger tiles up to the whole frame ($-11.9$\,dB) break what the decoder learned on its training window: per-tile global tokens, attention normalization and tile-normalized coordinates. Narrower overlaps keep the tiles but give weight to predictions near tile borders, which are less accurate ($-0.34$\,dB at 16\,px). Both act through attention; every other operation sees one token at a time. \method therefore keeps the tiles and their overlap but shares the tokens: each tile's attention sees exactly its released tokens, coordinates and global tokens; every other operation runs once per token; and an overlap token's attention outputs from its tiles are blended with the released weights.

Because the approximation enters at one kind of step, we can locate where it matters (\cref{sec:mechanism}). In MiniMax-H3, sharing the MLP is nearly free although it holds most of the computation. The cost arises when attention reads blended states and is concentrated in the first few layers, where a token's state still shapes its neighbours through attention. Keeping these layers per tile (\splitk{$k$}) recovers the released decoder. The required depth is a property of each model and is measured once: 4 of 36 layers in MiniMax-H3, 3 of 24 in MAGI-1~\cite{magi1} and none in ViTok-v2.

Without training, \splitk{6} is equivalent to the released MiniMax-H3 decoder in PSNR, LPIPS and DISTS on DAVIS videos and decodes $1.38$--$1.51\times$ faster than batched and compiled tiling on two GPU types without custom kernels (\cref{sec:exp}). The UVG dataset~\cite{uvg} was held out until both settings were frozen. On it, \splitk{4} (2--3\% faster still) is equivalent in all three metrics, and \splitk{6} is equivalent in PSNR and DISTS while improving LPIPS. \method carries over to MAGI-1 and to the ViTok-v2 image autoencoder~\cite{vitok}, again with confirmations on unseen data (\cref{sec:exp:others}). \method also composes with FP8 MLP layers (up to $1.96\times$), whereas matching the speed by changing the tiling costs at least 0.3\,dB. Finally, brief training can remove the remaining per-tile layers (\cref{sec:learn}): self-distillation does so on MiniMax-H3 without a loss in PSNR, and a coordinate field trained while the network stays frozen brings fully shared CLIP-based open-vocabulary segmentation~\cite{clip,sclip} to parity with sliding windows.

Our contributions are threefold. (1) \method: a training-free reformulation of overlapping tiled ViT decoding that keeps each tile's attention, computes everything else once per token without custom kernels and states precisely what is exact. (2) An interventional analysis showing that the approximation acts through attention rather than the MLP (MiniMax-H3) and mainly in the first layers (MiniMax-H3 and MAGI-1). (3) A rigorous evaluation on three decoders: equivalence on development data and in prospectively frozen confirmations on unseen data, a unified timing protocol, analyses by motion and tile zone, and an applicability audit of 14 configurations.

%% file: sec/2_related.tex
\section{Related Work}
\label{sec:related}

\paragraph{Overlapping windows.}
Models run beyond their training size on overlapping windows: tiled diffusion fuses or merges views to generate coherent large images~\cite{multidiffusion,spotdiffusion,mad}, and training-free dense CLIP slides windows over the image~\cite{sclip,naclip,clearclip,proxyclip,trident}; native-resolution tokenizers instead code whole images~\cite{vitok,atoken}. Convolutions share overlapping windows exactly because their output depends only on a fixed neighborhood: a sliding window equals one convolution over the whole image~\cite{overfeat}, overlapping windows can reuse activations~\cite{dontthinktwice}, and caching makes chunked decoding of causal convolutional video VAEs exact~\cite{wfvae,wan}. Attention normalized over a window has no such property, and overlapping local attention~\cite{halonet,natten,swin,hat} avoids duplication only because the model is trained with it; our setting is post hoc.

\paragraph{Sharing computation across attention windows.}
For dense CLIP, cross-window attention in the last layer~\cite{ovstitcher,glaclip} adds rather than removes computation. DistriFusion~\cite{distrifusion} splits high-resolution diffusion into patches across GPUs and gives each patch its neighbours' activations from the previous denoising step, \ie, stale context. Closest to our operator, the Neighborhood Patch Attention of ScaleDiff~\cite{scalediff} runs non-attention layers once on the full latent and lets non-overlapping query patches attend to overlapping key neighborhoods; it changes each query's context and blends nothing. \method instead keeps every query's released tiles and blends their outputs. On MiniMax-H3, keeping only each token's dominant tile while sharing every layer costs 1.21\,dB against 0.29\,dB with blending (\cref{tab:interventions}).

\paragraph{Removing the tiling with training.}
SPAR~\cite{spar} distills a sliding-window teacher into a single-pass ViT and Win-Win~\cite{winwin} trains high-resolution ViTs from two windows. In concurrent work, WindowFold~\cite{windowfold} folds CLIP's sliding windows into one forward pass with LoRA self-distillation. These methods remove the tiling and change the model; \method keeps the released tiling and weights, and our optional training (\cref{sec:learn}) keeps tile-exact attention.

\paragraph{Efficient decoding.}
Compact decoders~\cite{dcae,lightvae} trade quality for speed, and streaming decoders reorganize computation over time~\cite{flashdecoder}. System-level work compiles, quantizes, batches and parallelizes tile decodes~\cite{fastvideo,solh3,trtllmh3,vllmomnih3,comfyh3} while still decoding every tile in full; batched and compiled tiling is our baseline, and FP8 composes with \method. Token merging~\cite{tome,eventful} skips computation for \emph{similar} tokens; \method shares the computation of the \emph{same} token across tiles.

%% file: sec/3_method.tex
\section{Share Tokens, Keep Tiles}
\label{sec:method}

\subsection{Tiled Decoding Is a Hidden Ensemble}
\label{sec:ensemble}
Let $\mathcal{U}$ be the set of latent tokens. The released tiling defines overlapping tiles $t=1,\dots,T$; tile $t$ covers tokens $S_t\subset\mathcal{U}$ and carries its own global tokens $G_t$ (registers, a CLS token, or none). Each tile is decoded independently as $\hat y_t=D(x_{S_t},G_t)$, and the outputs are blended with stitching weights $w_t\ge0$, $\sum_t w_t=1$. An overlap token is thus predicted by an ensemble of tiles, each with a different context, and the redundancy $R=\sum_t|S_t|/|\mathcal{U}|$ is the average ensemble size. An overlap token $i$ thus carries one hidden state per covering tile, its \emph{per-tile states} $h_{t,i}$. A pre-norm block computes $h\leftarrow h+\mathrm{Attn}(\mathrm{N}(h))$ and $h\leftarrow h+\mathrm{MLP}(\mathrm{N}(h))$; only $\mathrm{Attn}$ mixes tokens.

Our running example is the MiniMax-H3 VAE decoder~\cite{minimaxh3}: 36 pre-norm ViT blocks of width 2048 with SwiGLU MLPs and QK-RMSNorm. Positions enter through 3-axis RoPE~\cite{rope} on \emph{per-tile normalized} coordinates, and every tile carries 4 registers as its global tokens. The released tiling uses $16\times16$-token tiles (256\,px) whose overlap of at least 64\,px is enlarged until the tiles exactly cover the frame; the tiles are blended linearly in pixel space. Temporal windows hold 7 latents, overlap by 2 and are cross-faded. We also evaluate two decoders that differ from MiniMax-H3 in position encoding, global tokens or domain. MAGI-1~\cite{magi1} is a video decoder of 24 blocks with absolute position embeddings and one CLS token per tile; it is released with 256\,px tiles, 25\% overlap and 12-frame temporal chunks. ViTok-v2~\cite{vitok} is an image autoencoder whose decoder has 24 parallel blocks (attention and MLP read the same normalized input), 2D RoPE on integer coordinates and no global tokens. It is released for whole-image coding; at 2K--4K, we tile its encoder and decoder with 256\,px tiles and 25\% overlap (\cref{sec:exp:others}).

This ensemble is expensive. Transformer blocks take over 98\% of the decoder's GPU time and attention only 9.6\%; the rest is per-token normalization, projections and MLPs. Spatial overlap gives $R=2.03$ at 704$\times$1280 and $1.77$ at 768$\times$1344; hence 44--51\% of the per-token computation re-processes tokens that another tile also processes.

\subsection{Why Tiles Must Be Kept}
\label{sec:tile}
The obvious ways to remove the duplication all lose quality (\cref{fig:pareto}, supp.~I). Narrower overlaps shift weight to predictions near tile borders, which are less accurate; larger tiles break three properties that the decoder learned from single tiles.
(i) \emph{Global tokens are tile-bound}~\cite{registers}: dropping MiniMax-H3's registers costs 8--9\,dB, and registers computed for windows other than the released tiles leave seams.
(ii) \emph{Attention normalization is tile-bound}: quality falls monotonically once a query sees more keys than in training, down to $-12.5$\,dB for full-frame attention.
(iii) \emph{Coordinates are tile-bound}: a token has different RoPE coordinates in different tiles, and giving the registers other coordinates than in training costs 5--7\,dB.
The best single-pass sliding window we found still loses 0.67\,dB and is no faster than batched tiling. Two conclusions follow. Each tile's attention must see exactly its released tokens, coordinates and global tokens. None of these failures involves the per-token operations, which never look at the tile yet hold almost all of the duplicated computation.

\subsection{\method}
\label{sec:tileshare}
\begin{figure}[t]
\centering
\includegraphics[width=\linewidth]{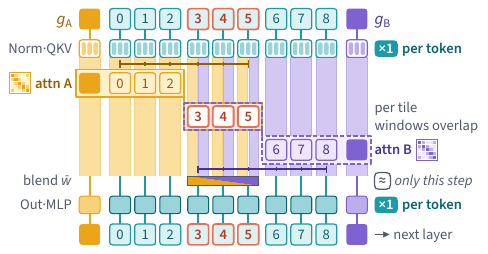}
\caption{\textbf{Inside one shared layer.} Each token has one state: normalization, QKV projections and QK-normalization run once per token (teal), as do the out-projection and MLP later. Attention still runs per released tile, with the tile's own global token $g$ and coordinates; rulers mark each window's extent. The overlap tokens 3--5 are stored once but attended in both windows, and their two attention outputs are blended with the released weights $\bar w$, the only approximate step.}
\label{fig:layer}
\end{figure}

\method keeps one hidden state $h_i$ per unique token $i\in\mathcal{U}$, plus the global tokens of every tile (\cref{fig:layer}). In each block, normalization, QKV projections and QK-normalization are computed once per unique token. For every tile $t$ we then gather $\{q_i,k_i,v_i\}_{i\in S_t}$ with the tile's own global tokens, apply the positional encoding with the tile's own coordinates exactly as the released decoder does, and run dense attention; all tiles share one batched attention call. The tiles' outputs for an overlap token are blended as $\tilde a_i=\sum_{t\ni i}\bar w_{t,i}\,a_{t,i}$ with the released stitching weight pooled to the token; global tokens keep their own tile's output. Out-projection, residual and MLP are again computed once per token.

\paragraph{What is exact.}
In decoders whose positions enter only through attention (\eg, MiniMax-H3 and ViTok-v2), a token enters the first layer with the same state in every tile that covers it. Computing the operations before attention once is therefore exact. With the released call shapes, the first layer of MiniMax-H3 is bit-identical to the released decoder in its normalization, QKV projections, QK-normalization and per-tile attention outputs; so is the whole first layer for tokens covered by a single tile. Our batched implementation issues other call shapes and matches up to fp16 rounding (relative $\ell_2$ $3\cdot10^{-4}$), the same order as between the released decoder's own per-tile and batched calls. The approximation enters at the blend: an overlap token's per-tile attention outputs become one value; from then on, attention reads a shared state where the baseline keeps one state per tile. In decoders that add absolute position embeddings at the input (\eg, MAGI-1), a token's per-tile states differ from the start. Computing even the first layer's per-token operations once is therefore approximate. Supplementary~A lists every operation and verifies the implementation on a tiling without overlap.

\paragraph{Per-tile depth and cost.}
In \splitk{$k$} (per-tile depth $k$), layers $0,\dots,k-1$ run per tile exactly as in the baseline. At layer $k$, the per-tile states are merged as $h_i=\sum_t\bar w_{t,i}h_{t,i}$, and \method runs the remaining layers (\cref{fig:teaser}, right). This merge is approximate as well, because the per-tile states have diverged by then. \splitk{0} shares every layer. At the other end, \splitk{$L$} ($L$ blocks) runs every block per tile and merges the per-tile states only before the output head. It differs from the released decoder only in blending token states with one weight per token instead of pixels, which costs 0.008\,dB. Linear-layer FLOPs relative to the baseline are $(kR+L-k)/(LR)$, and attention FLOPs are unchanged. Let $\alpha$ be the attention share of the baseline's FLOPs. A cost model proportional to FLOPs then bounds the speed-up of \splitk{0} by $1/(\alpha+(1-\alpha)/R)$: $1.84\times$ for MiniMax-H3 ($\alpha=0.10$, $R=2.03$; measured: $1.69\times$) and $1.38\times$ for MAGI-1 ($\alpha\approx0.30$, $R=1.65$).
For MiniMax-H3, one sweep over a few depths on eight development clips (Dev-8) gave $k=4$ without labels, since an autoencoder's ground truth is its input. The sweep on All-38 (Dev-8 plus 30 DAVIS videos; \cref{sec:exp:setup}) shows a plateau at $k=4$--$6$ (\cref{sec:mech:depth}). We report both ends: the faster \splitk{4} and \splitk{6}, which also keeps LPIPS and DISTS within their margins. Because \splitk{6} was chosen on this sweep, which includes the DAVIS evaluation clips, both settings were frozen before any evaluation on unseen data (\cref{sec:exp:main}). The small temporal overlap ($7/5$ windows per latent) stays per window (supp.~B).

\paragraph{Implementation.}
Gather indices and per-tile RoPE tables are precomputed per latent shape, blends run in fp32, and every layer is compiled with \ttt{torch.compile} without custom kernels. On MiniMax-H3, peak memory stays within 2\,GB of batched tiling.

%% file: sec/4_mechanism.tex
\section{Where the Approximation Matters}
\label{sec:mechanism}

\begin{table}[t]
\centering\small
\setlength{\tabcolsep}{3.5pt}
\caption{\textbf{Where the approximation acts} (MiniMax-H3, All-38; $\Delta$PSNR in dB relative to the released decoder). \emph{Top:} from layer $k$ on, each tile keeps its own residual stream and only the listed operations are shared; attention always runs per tile. FLOPs: block FLOPs relative to the baseline at 704p. $^\ast$Equal to \splitk{0}, since identical per-tile states stay identical. \emph{Bottom:} rule for blending an overlap token's tiles.}
\label{tab:interventions}
\resizebox{\linewidth}{!}{\begin{tabular}{lcccc}
\toprule
& \multicolumn{2}{c}{$k=4$} & \multicolumn{2}{c}{$k=0$} \\
\cmidrule(lr){2-3}\cmidrule(lr){4-5}
Shared operations & $\Delta$PSNR & FLOPs & $\Delta$PSNR & FLOPs \\
\midrule
none (per-tile states throughout) & $-0.008$ & 1.000 & $-0.008$ & 1.000 \\
QKV only & $-0.070$ & 0.924 & $-3.262$ & 0.914 \\
out-projection only & $-0.785$ & 0.975 & $-0.290^\ast$ & 0.971 \\
MLP only & $-0.016$ & 0.695 & $-0.022$ & 0.657 \\
all (\method) & $-0.003$ & 0.593 & $-0.290$ & 0.543 \\
\midrule
Blending rule & \multicolumn{2}{c}{$k=4$} & \multicolumn{2}{c}{$k=0$} \\
\midrule
linear ramp (released) & \multicolumn{2}{c}{$\mathbf{-0.003}$} & \multicolumn{2}{c}{$\mathbf{-0.290}$} \\
Gaussian, $\sigma=0.25$ tile & \multicolumn{2}{c}{$-0.094$} & \multicolumn{2}{c}{$-0.370$} \\
dominant tile only & \multicolumn{2}{c}{$-0.298$} & \multicolumn{2}{c}{$-1.214$} \\
uniform average & \multicolumn{2}{c}{$-0.821$} & \multicolumn{2}{c}{$-1.494$} \\
\bottomrule
\end{tabular}}
\end{table}

\subsection{Through attention, not through the shared MLP}
\label{sec:mech:path}
Computing the MLP once is nearly free; the cost arises where attention reads blended states. To separate the paths, we let each tile keep its own residual stream as in the baseline and share only selected operations from layer $k$ on; everything else runs per tile (\cref{tab:interventions}, top): QKV, computed once on the weighted average of the per-tile states; the out-projection, applied once to the blended attention outputs and added to every tile's stream; or the MLP, computed once on the weighted average and added to every tile's stream. Sharing the MLP alone yields three quarters of the FLOP savings of full sharing and is nearly free even from the first layer ($-0.022$\,dB). Sharing QKV and MLP already equals full \method at any $k$, because the weighted average of the per-tile states then follows the shared state exactly. In MiniMax-H3, the loss of full sharing ($-0.29$\,dB at $k=0$) thus appears once QKV is computed from blended states as well. Sharing only the MLP would need no $k$, but it runs the attention path on every tile's stream and is 6--7\% slower than \splitk{6} (supp.~F).

\subsection{Mainly in the first few layers}
\label{sec:mech:depth}
\begin{figure}[t]
\centering
\includegraphics[width=\linewidth]{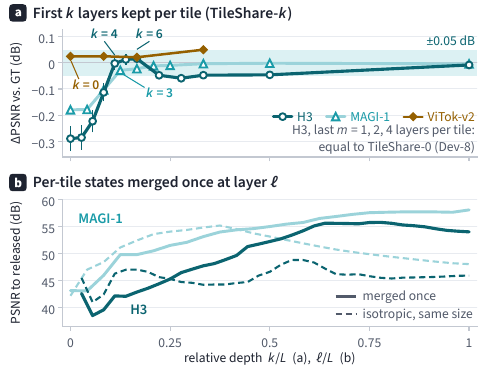}
\caption{\textbf{Depth.} (a) $\Delta$PSNR relative to tiled decoding for \splitk{$k$} against $k/L$ (MiniMax-H3: All-38, 95\% CIs; MAGI-1: DAVIS-10; ViTok-v2 350M: DIV8K-40 at $4096^2$, CIs within the markers); shaded: $\pm0.05$\,dB. (b) Merging the per-tile states once at layer $\ell$, then continuing per tile: PSNR of the output to the released output (solid), and for an isotropic perturbation of the same size (dashed); means over 30 (MiniMax-H3) and 42 (MAGI-1) temporal windows.}
\label{fig:mechanism}
\end{figure}

Keeping the first layers per tile recovers the released decoder, whereas keeping the last 1--4 does not reduce the loss of \splitk{0} (\cref{fig:mechanism}a). In MiniMax-H3, three per-tile layers recover 61\% of the loss of \splitk{0} ($-0.290\to-0.113$\,dB) and four recover all of it ($-0.003$\,dB); $k=4$--$6$ form a plateau of equivalence. MAGI-1 differs in position encoding, global tokens and depth (absolute instead of rotary positions, one CLS token instead of four registers, 24 instead of 36 layers) and reaches the plateau at $k=3$. ViTok-v2 needs no per-tile layer at all (\cref{sec:exp:others}). Between $k=8$ and 18, MiniMax-H3 trades about 0.05\,dB of PSNR for a lower LPIPS (supp.~E); we use the plateau.

Merging the per-tile states once at a single layer locates where this happens, without ground truth (\cref{fig:mechanism}b). In both decoders, its cost peaks in the first layers and falls with depth; summed from layer $k$ on, it ranks the deviation of \splitk{$k$} from the released decoder exactly (Spearman 1.0). Two observations indicate why the cost falls. In MiniMax-H3, swapping one tile's overlap states for its neighbour's changes the tile's \emph{non-overlap} pixels substantially at layer 1 (PSNR 39\,dB to the original) but barely at layer 34 (84\,dB): early states shape their neighbours through attention, whereas late states mostly shape their own output. Moreover, a merge costs up to 5\,dB more than an isotropic perturbation of the same size in layers 1--8 but 3.5--10\,dB less from mid-depth on. Early per-tile states thus differ in directions to which the output is sensitive, and later ones in directions it ignores.

The required depth $k^\star$ is model-specific, from 0 to all layers (\cref{sec:exp:scope}); we measure it once per model and deployment geometry, on a few development clips (diagnostics without ground truth: supp.~E).

\subsection{What blending buys, and which blend}
\label{sec:mech:blend}
The released blending beats hard stitching (each pixel from its dominant tile) by 0.19\,dB, almost entirely in overlap bands. There, two tiles' errors correlate ($\rho=0.69$--$0.90$), and a single tile's error grows towards its border, up to $1.64\times$ the blended error; the ramp gives these border predictions little weight. \method inherits this benefit only if it weights tiles the same way (\cref{tab:interventions}, bottom). The released linear ramp pooled to tokens is best. At \splitk{4}, a Gaussian loses 0.09\,dB and keeping only each token's dominant tile 0.30\,dB; uniform averaging, which trusts a tile's border as much as its center, loses 0.82\,dB.

%% file: sec/5_experiments.tex
\section{Experiments}
\label{sec:exp}

\begin{table*}[t]
\centering\small
\setlength{\tabcolsep}{4pt}
\caption{\textbf{MiniMax-H3 without training.} $\Delta$PSNR (dB) vs.\ ground truth relative to the released tiled decoder [90\% CI]. DAVIS-30: held out from the choice of $k=4$, not of $k=6$. UVG: evaluated once with both settings frozen. P\,/\,L\,/\,D (UVG: both resolutions): \checkmark{} equivalent, the 90\% CI within the margin ($\pm0.05$\,dB; $\pm1.16\%$ of the baseline mean for LPIPS, DISTS); $\uparrow$ better, the CI leaving the margin only on the side of improvement; $\times$ neither; -- not measured. Speed-up of the compiled decoder alone on an RTX PRO 6000 vs.\ batched tiling at 704p\,/\,768p\,/\,1080p.}
\label{tab:main}
\resizebox{\textwidth}{!}{\begin{tabular}{lccccccc}
\toprule
& \multicolumn{2}{c}{DAVIS-30} & \multicolumn{3}{c}{UVG (prospective)} & \\
\cmidrule(lr){2-3}\cmidrule(lr){4-6}
Method & $\Delta$PSNR & P\,/\,L\,/\,D & $\Delta$PSNR, 704p & $\Delta$PSNR, 1080p & P\,/\,L\,/\,D & Speed-up \\
\midrule
Released tiling (batched) & 0 & & 0 & 0 & & 1.00 / 1.00 / 1.00 \\
Hard stitching & $-0.199$ [$-0.212,-0.186$] & $\times$\,/\,--\,/\,-- & -- & -- & & -- \\
\midrule
\splitk{0} & $-0.302$ [$-0.343,-0.260$] & $\times$\,/\,$\times$\,/\,$\times$ & -- & -- & & 1.69 / 1.54 / 1.61 \\
\splitk{4} & $-0.011$ [$-0.030,+0.009$] & \checkmark\,/\,$\times$\,/\,$\times$ & $-0.006$ [$-0.031,+0.018$] & $-0.002$ [$-0.027,+0.024$] & \checkmark\,/\,\checkmark\,/\,\checkmark & 1.56 / 1.44 / 1.50 \\
\splitk{6} & $\mathbf{+0.009}$ [$-0.006,+0.025$] & \checkmark\,/\,\checkmark\,/\,\checkmark & $+0.018$ [$+0.000,+0.036$] & $+0.031$ [$+0.013,+0.049$] & \checkmark\,/\,$\uparrow$\,/\,\checkmark & 1.51 / 1.41 / 1.45 \\
\midrule
FP8 MLP layers only & $-0.008$ [$-0.013,-0.004$] & \checkmark\,/\,\checkmark\,/\,-- & -- & -- & & 1.35 / 1.35 / 1.35 \\
\splitk{4} + FP8 & $-0.020$ [$-0.040,+0.001$] & \checkmark\,/\,$\times$\,/\,$\times$ & -- & -- & & 2.04 / 1.88 / 1.94 \\
\splitk{6} + FP8 & $+0.000$ [$-0.016,+0.017$] & \checkmark\,/\,\checkmark\,/\,\checkmark & -- & -- & & 1.96 / 1.85 / 1.90 \\
\bottomrule
\end{tabular}}
\end{table*}

\subsection{Setup}
\label{sec:exp:setup}
We evaluate MiniMax-H3 on two independent sources of real video. DAVIS-30 consists of 30 DAVIS 2017~\cite{davis} sequences (sources 720p--4K) resized and center-cropped to 704$\times$1280, 73 frames each; 1080p-26 re-evaluates at 1088$\times$1920 the 26 of them whose sources reach 1080p. Eight further clips (Dev-8: five generated and three real) were used while designing the method, including the sweep that gave $k=4$; All-38 is their union with DAVIS-30. For a prospective confirmation, we froze \splitk{4} and \splitk{6} and only then evaluated them at 704$\times$1280 and 1088$\times$1920 on all 16 sequences of the previously unused UVG dataset~\cite{uvg}. Gen-44 are 44 videos from the 4-step FastH3 pipeline at 768$\times$1344$\times$124; for lack of ground truth, they are compared with the released decoder. MAGI-1 is evaluated on the 10 DAVIS videos on which we chose $k$ and once on the 16 UVG sequences at 704p. ViTok-v2 is evaluated on $2048^2$ and $4096^2$ crops of the 40 DIV8K images~\cite{div8k} on which we chose $k$ and once on 40 unseen Wikimedia Commons photographs (Commons-40; supp.~K).

Quality is judged by equivalence to the released decoder. PSNR, LPIPS~\cite{lpips} and DISTS~\cite{dists} against ground truth are reported as differences to the released tiled decoder, with paired bootstrap CIs over videos. A variant is \emph{equivalent} if two one-sided tests (TOST) at level 0.05 place the mean difference within the margin (\ie, its 90\% CI lies inside it). It is \emph{non-inferior} if its 95\% CI excludes a loss larger than the margin (a lower PSNR or a higher LPIPS or DISTS). The PSNR margin of $\pm0.05$\,dB is about a quarter of what blending buys (\cref{sec:mech:blend}). The LPIPS and DISTS margins are 1.16\% of the baseline's mean, which is the relative change in squared error that corresponds to 0.05\,dB.

We time the decoder alone, from latents to blended pixels on the GPU, at release precision (fp32 weights and fp16 autocast). Every method is compiled with \ttt{torch.compile}, and we report the median of ten runs after compilation and two warm-ups on an RTX PRO 6000 with no other active process. The baseline batches all tiles of a temporal window into one call; in GPU time this is $1.4\times$ faster than per-tile calls, which would otherwise inflate our gains. ViTok-v2 is timed the same way in bf16; supp.~C adds H100 timings, batching, memory and the end-to-end pipeline.

\subsection{Main results on MiniMax-H3}
\label{sec:exp:main}
\splitk{6} is equivalent to the released decoder in PSNR, LPIPS and DISTS on DAVIS-30, and the prospective UVG confirmation supports it (\cref{tab:main}): there it is equivalent in PSNR and DISTS at both resolutions and lowers LPIPS. Chosen on Dev-8 alone, \splitk{4} is equivalent in all three metrics on UVG and in PSNR on DAVIS-30, where its LPIPS and DISTS change by 1.6\% of the baseline. All-38 and 1080p-26 agree (supp.~B).

The speed-up holds on two GPU types and comes from the per-token GEMMs. \splitk{6} decodes $1.41$--$1.51\times$ faster than batched and compiled tiling and \splitk{4} $1.44$--$1.56\times$; on an H100 the two reach $1.38$--$1.50\times$ and $1.42$--$1.55\times$. At 704p, \splitk{4} saves 3.14\,s of GPU time per clip (2.08\,s in the MLP GEMMs), while attention still runs per tile at unchanged cost. The measured speed-ups of \splitk{4} and \splitk{6} reach 93--95\% of their FLOP bounds at all three resolutions; the rest is mostly new gather and blend traffic (supp.~C).

Generated videos and downstream models tell the same story. On Gen-44, \splitk{6} reaches a PSNR of 42.7\,dB to the released decoder's output (\splitk{4}: 41.5\,dB), and four VBench~\cite{vbench} scores change negligibly. Depth~\cite{depthanythingv2}, segmentation~\cite{segformer} and optical-flow~\cite{raft} models applied to the decoded frames change their predictions by only 29--37\% of the change caused by the released decoder's own error (\splitk{4}; \splitk{6} less; supp.~H). End to end, \splitk{4} speeds up a single-pass request by $1.08\times$ ($1.11\times$ with FP8) and shortens the decoding stage of a 4-step request from 6.45 to 4.54\,s (supp.~C).

\subsection{Against alternatives}
\label{sec:exp:alt}
\begin{figure}[t]
\centering
\includegraphics[width=\linewidth]{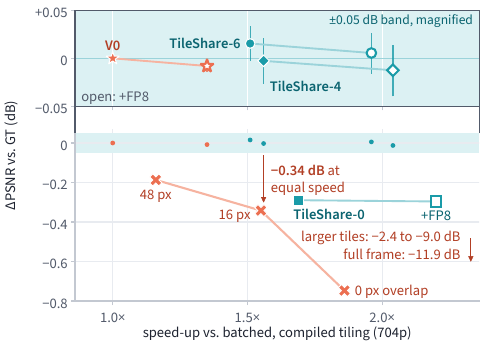}
\caption{\textbf{Quality vs.\ decoder speed-up} on MiniMax-H3 (All-38, 704p; larger tiles and whole-frame decoding on 18 clips). Top: the $\pm0.05$\,dB band, magnified on the same speed axis; open markers add FP8 MLP layers; bars are 95\% CIs.}
\label{fig:pareto}
\end{figure}

Changing the tiling cannot match this trade-off (\cref{fig:pareto}). Shrinking the overlap to 16\,px is as fast as \splitk{4} at 704p ($1.55\times$) but loses 0.34\,dB; a minimum overlap of 0\,px reaches $1.86\times$ at $-0.75$\,dB with visible seams; and larger tiles or whole-frame decoding lose 2.4--11.9\,dB. Other accelerators sit elsewhere: the tiny decoder TAEH3 is 22--31$\times$ faster but loses 5.5\,dB (LPIPS $+0.10$); W8A8 INT8 keeps quality but gives only $1.02\times$ on our GPU; and FP8 MLP layers keep quality at $1.35\times$ and compose with \method.

\subsection{Beneath the mean}
\label{sec:exp:beneath}
\begin{table}[t]
\centering\small
\setlength{\tabcolsep}{3pt}
\caption{\textbf{Beneath the mean.} Fastest third: $\Delta$PSNR (dB) [95\% CI] on the third of All-38 with the largest optical flow. Corners: four-tile corners of 1080p-26. Steps: net excess luminance steps of more than 6 gray levels across tile-start lines, per mille of 32-px segments (Gen-44\,/\,1080p-26\,/\,UVG 1080p; supp.~B). HF: high-frequency energy relative to the released decoder (All-38\,/\,UVG 1080p).}
\label{tab:beneath}
\resizebox{\linewidth}{!}{\begin{tabular}{lcccc}
\toprule
& Fastest third (All-38) & Corners (1080p) & Steps (\textperthousand) & HF (\%) \\
\midrule
\splitk{4} & $-0.051$ [$-0.077,-0.023$] & $-0.162$ & 4.8 / 5.7 / 2.6 & $-4.7$ / $-3.1$ \\
\splitk{6} & $-0.030$ [$-0.047,-0.011$] & $-0.107$ & 2.0 / 1.3 / 0.7 & $-2.6$ / $-1.6$ \\
\bottomrule
\end{tabular}}
\end{table}

\begin{figure}[t]
\centering
\includegraphics[width=\linewidth]{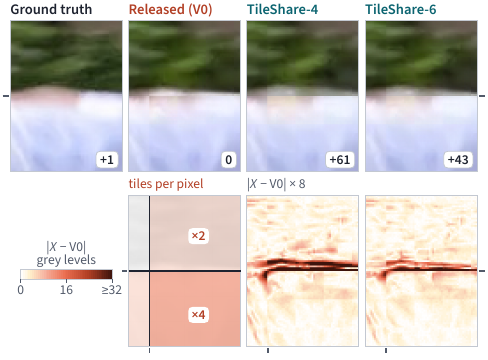}
\caption{\textbf{The largest local step} among held-out real clips, selected on \splitk{4} alone (1080p soapbox, frame 65, start row of a four-tile corner). Top: 80$\times$80\,px crops of ground truth, the released decoder, \splitk{4} and \splitk{6}, with the excess step over the released decoder in gray levels. Bottom: tiles per pixel and the absolute difference to the released output ($\times8$); ticks mark tile-start lines.}
\label{fig:seams}
\end{figure}

Below the mean, the remaining differences concentrate in three places, each reduced by \splitk{6} (\cref{tab:beneath}). \emph{Motion:} losses grow with motion; per-video differences correlate with flow magnitude (Spearman $-0.5$ to $-0.6$). \emph{Tile borders:} \splitk{4} shows excess luminance steps on 2--6\textperthousand{} of tile-start segments, nearly all inside four-tile corners (89--98\%), where the released stitching itself switches abruptly between tiles. The steps are local and last a single frame; \splitk{6} removes 57--76\% of them, but not the largest (\cref{fig:seams}). \emph{Texture:} both settings are slightly smoother than the baseline in high-frequency energy. Where fast motion or fine texture dominate, \splitk{6} is the recommended setting.

\subsection{Two further decoders}
\label{sec:exp:others}
\begin{table}[t]
\centering\small
\setlength{\tabcolsep}{2.5pt}
\caption{\textbf{MAGI-1 and ViTok-v2 without training.} $\Delta$PSNR (dB) relative to overlapping tiled decoding [90\% CI]; P\,/\,L\,/\,D as in \cref{tab:main}. For each decoder, the first set chose $k$ and the second was evaluated once with $k$ frozen. MAGI-1: released setting, UVG at 704p; ViTok-v2 350M: $2048^2$\,/\,$4096^2$. Whole: whole-input coding relative to tiled coding. Speed-up: decoder only, vs.\ batched tiling\,/\,vs.\ whole-input decoding (MAGI-1: 704p\,/\,720p). $^\dagger$On eight other clips.}
\label{tab:others}
\resizebox{\linewidth}{!}{\begin{tabular}{llcccc}
\toprule
Decoder, data & $k$ & $\Delta$PSNR & P\,/\,L\,/\,D & Whole & Speed-up \\
\midrule
MAGI-1, DAVIS-10 & 3 & $-0.029$ [$-0.035,-0.023$] & \checkmark\,/\,\checkmark\,/\,\checkmark & $-4.1^\dagger$ & $1.13$\,/\,$1.14$ \\
MAGI-1, UVG-16 & 3 & $-0.027$ [$-0.032,-0.023$] & \checkmark\,/\,$\times$\,/\,\checkmark & -- & -- \\
\midrule
ViTok-v2, DIV8K-40 & 0 & $+0.023$ [$+0.014,+0.031$] & \checkmark\,/\,$\uparrow$\,/\,$\uparrow$ & $-0.95$ & $1.77$\,/\,$2.54$ \\
& & $+0.024$ [$+0.016,+0.033$] & \checkmark\,/\,$\uparrow$\,/\,$\uparrow$ & $-2.00$ & $1.52$\,/\,$7.35$ \\
ViTok-v2, Commons-40 & 0 & $+0.026$ [$+0.017,+0.037$] & \checkmark\,/\,$\uparrow$\,/\,$\uparrow$ & $-1.25$ & -- \\
& & $+0.024$ [$+0.014,+0.033$] & \checkmark\,/\,$\uparrow$\,/\,$\uparrow$ & $-2.03$ & -- \\
\bottomrule
\end{tabular}}
\end{table}

\method carries over to two further decoders (\cref{tab:others}). On MAGI-1 with its released 12-frame chunks and per-tile output head, \splitk{3} is equivalent to tiled decoding in PSNR, LPIPS and DISTS on DAVIS-10, where we chose $k$. On all 16 UVG sequences, evaluated once with $k$ frozen, it is equivalent in PSNR and DISTS, and its mean LPIPS change ($+0.0014$) also lies within the margin. It decodes $1.13$--$1.14\times$ faster than batched tiling, 86--87\% of its cost bound ($1.30$--$1.32\times$), which is set by attention's 30\% share of MAGI-1's FLOPs. Keeping a whole clip's token states resident raises peak memory by 5--9\,GB (supp.~K).

ViTok-v2 is released for whole-image coding at any resolution. Yet at $2048^2$ and $4096^2$, coding the image in overlapping 256\,px tiles is 1--2\,dB better, and whole-image coding shows a 16\,px token grid (supp.~K). Non-overlapping tiles are faster but leave a discontinuity at every tile border; overlapping tiles avoid it, and \method recovers 80--88\% of the speed difference (supp.~K). \method needs no per-tile layer here: on DIV8K-40, each tested $k\in\{0,2,4\}$ is equivalent in PSNR, and we froze $k=0$. On Commons-40, evaluated once afterwards, \splitk{0} is again equivalent in PSNR and better than tiled decoding in LPIPS and DISTS by more than the margins. It decodes $1.52$--$1.77\times$ faster than batched tiling, close to its FLOP bound ($1.67$--$1.83\times$), with less peak memory. Results for the 5B model are in supp.~K.

\subsection{Applicability and limits}
\label{sec:exp:scope}
\begin{table}[t]
\centering\small
\setlength{\tabcolsep}{3pt}
\caption{\textbf{Fourteen configurations.} Whole $-$ tiled: whole-input inference relative to tiled inference (dB PSNR for decoders, mIoU for encoders). $k^\star$: smallest per-tile depth that meets the criterion of its model family without training (supp.~F). Speed-up of \splitk{$k^\star$} vs.\ batched tiling; -- none or not measured. $^\dagger$Inconclusive; better whole at short side 896. $^\ddagger$$4.1\times$ slower than tiled decoding with \method at 4K. $^\S$Batch 8 ($0.89\times$ at batch 1). $^\P$See \cref{sec:exp:others}; ViTok-v2 at $2048^2$--$4096^2$.}
\label{tab:scope}
\resizebox{\linewidth}{!}{\begin{tabular}{llccc}
\toprule
Model & Positions & Whole $-$ tiled & $k^\star/L$ & Speed-up \\
\midrule
\multicolumn{5}{l}{\emph{Whole input as good in quality}} \\
DINOv3-B/16 (enc.)~\cite{dinov3} & RoPE & $-0.09^\dagger$ & 8/12 & -- \\
DINOv3-L/16 (enc.) & RoPE & $-0.11^\dagger$ & 16/24 & -- \\
PE-Spatial-B/16 (enc.)~\cite{perceptionencoder} & RoPE + abs. & $+0.13$ & 0/12 & 1.32 \\
ViTok-v2 5B (image dec.)$^\P$~\cite{vitok} & RoPE & $+0.36^\ddagger$ & 0/40 & 1.61 \\
\midrule
\multicolumn{5}{l}{\emph{Tiling needed, small $k^\star$}} \\
MiniMax-H3 (video dec.) & RoPE & $-11.9$ & 4/36 & 1.44--1.56 \\
MAGI-1 (video dec.)$^\P$~\cite{magi1} & abs. & $-4.1$ & 3/24 & 1.13--1.14 \\
ViTok-v2 350M (image dec.)$^\P$ & RoPE & $-0.9$ & 0/24 & 1.52--1.77 \\
RAE-MAE-B (image dec.)~\cite{rae} & none & $-10.2$ & 0/28 & 1.40 \\
SBB ViT-B/16 (enc.)~\cite{timm} & RoPE & $-1.0$ & 0/12 & 1.20$^\S$ \\
\midrule
\multicolumn{5}{l}{\emph{Tiling needed, deep $k^\star$}} \\
RAE-SigLIP2-B (image dec.) & abs. & $-4.2$ & 12/28 & 1.19 \\
RAE-DINOv2-B (image dec.) & abs. & $-5.2$ & 16/28 & 1.13 \\
RAE-DINOv2-L (image dec.) & abs. & $-6.0$ & 14/28 & 1.13 \\
DINOv2-B/14 (enc.)~\cite{dinov2} & abs. & $-0.5$ & 12/12 & -- \\
CLIP ViT-B/16, SCLIP (enc.)~\cite{clip,sclip} & abs. & $-2.4$ & 12/12 & -- \\
\bottomrule
\end{tabular}}
\end{table}

Of 14 configurations, four need no tiling for quality; of the ten that do, five are served at small $k^\star$ and five need a large per-tile depth (\cref{tab:scope}). This exploratory audit applies the same procedure to two video decoders, six image decoders and six window-bound encoders, building the tiled pipeline ourselves where the model does not ship one (supp.~F). Two measurements decide applicability. A whole-input test first asks whether tiling is needed at all. One pass over the whole input is as good for PE-Spatial and ViTok-v2 5B and for DINOv3 at high resolution, although whole-image decoding with ViTok-v2 5B is $4.1\times$ slower at 4K; elsewhere it loses 0.9--11.9\,dB or 0.5--2.4 mIoU. Second, $k^\star$ is small for MiniMax-H3, MAGI-1, ViTok-v2 350M, the MAE decoder of RAE~\cite{rae} and the SBB encoder, whereas the RAE decoders of SigLIP2 or DINOv2 features need 12--16 of 28 layers and the DINOv2 and CLIP encoders every layer; for encoders, $k^\star$ also grows with resolution (supp.~F).

%% file: sec/6_learn.tex
\section{Sharing Every Layer with Training}
\label{sec:learn}

\begin{table}[t]
\centering\small
\setlength{\tabcolsep}{2.5pt}
\caption{\textbf{Sharing every layer with training} (Trained: new parameters; the coordinate field leaves the network frozen). MiniMax-H3: $\Delta$PSNR (dB) relative to the released decoder; worst coverage zone; decoder speed-up at 704p. CLIP: $\Delta$ of the mean mIoU over seven benchmarks relative to sliding windows; worst benchmark; backbone speed-up.}
\label{tab:learn}
\resizebox{\linewidth}{!}{\begin{tabular}{lcccc}
\toprule
Variant & Trained & $\Delta$ [95\% CI] & Worst & Speed-up \\
\midrule
\multicolumn{5}{l}{\emph{MiniMax-H3, All-38}} \\
\splitk{0} & -- & $-0.290$ [$-0.337,-0.242$] & $-0.54$ & 1.69 \\
\splitk{0} + LoRA self-distillation & 51.9M & $\mathbf{+0.008}$ [$-0.017,+0.032$] & $\mathbf{-0.01}$ & 1.69 \\
\splitk{0} + coordinate field & 1.3M & $-0.086$ [$-0.122,-0.050$] & $-0.17$ & -- \\
\midrule
\multicolumn{5}{l}{\emph{CLIP open-vocabulary segmentation, mean over seven benchmarks}} \\
SCLIP, \splitk{0} & -- & $-2.03$ [$-2.29,-1.78$] & -- & 1.71--2.15 \\
SCLIP, \splitk{0} + coordinate field & 0.64M & $+0.17$ [$+0.03,+0.29$] & $-0.08$ & 1.71--2.15 \\
NACLIP, \splitk{0} + same field & -- & $+0.10$ [$-0.04,+0.21$] & $-0.18$ & -- \\
ClearCLIP, \splitk{0} + same field & -- & $+0.05$ [$-0.13,+0.22$] & $-0.59$ & -- \\
\bottomrule
\end{tabular}}
\end{table}

ViTok-v2 already shares every layer without training; MiniMax-H3 and MAGI-1, trained only on independent tiles, have never seen shared states. Brief training recovers most or all of the loss of sharing every layer; unlike the rest of the paper, this section requires it (\cref{tab:learn}).

\paragraph{Self-distillation.}
The teacher is the frozen released decoder with its released tiling. The student is the same decoder run as \splitk{0} with rank-32 LoRA~\cite{lora} adapters on all linear layers. It is trained for 2500 steps (about two GPU-hours) with pixel L1 plus $0.1\cdot$LPIPS between the two outputs on random crops of $3\times3$ to $4\times4$ tiles from 60 DAVIS sequences and 30 generated latents that are disjoint from all test data; no ground truth is needed. The student is equivalent to the released decoder in PSNR, also with a second seed and at 1080p. Its LPIPS and DISTS change by about 1.3\% of the baseline ($+0.0010$ and $+0.0006$ on All-38) as for \splitk{4}, and it runs at the speed of \splitk{0}. Every coverage zone of All-38 stays within 0.04\,dB of the baseline, while the four-tile corners at 1080p reach $-0.08$\,dB [$-0.13$, $-0.04$], half the loss of \splitk{4}. Applied to a tiling with less overlap, the same recipe recovers only half of the loss and leaves visible seams at 0\,px, whereas the \method student sees exactly its teacher's context. Tile-exact attention thus matters even with training.

\paragraph{A coordinate field on a frozen decoder.}
Training can also leave every decoder weight untouched: a small network maps Fourier features of each token's global coordinates to a residual offset per layer, injected through zero-initialized gates, trained by the same self-distillation and precomputed per latent shape. With 1.3M parameters it recovers 70\% of the PSNR gap of \splitk{0} on MiniMax-H3; with 1.2M parameters on MAGI-1, it recovers 75\%, more than LoRA (61\%).

\paragraph{Encoders.}
A coordinate field also makes CLIP, which needs every layer per window without training (\cref{tab:scope}), fully shared. One field of 0.64M parameters trained with SCLIP~\cite{sclip} as the teacher serves SCLIP, NACLIP~\cite{naclip} and ClearCLIP~\cite{clearclip}, the last with rescaled coordinates. At $k=0$, their mean mIoU over seven open-vocabulary segmentation benchmarks matches sliding windows (supp.~G). The backbone runs $1.71$--$2.15\times$ faster; end to end, a single image is $1.17\times$ faster, because upsampling the class logits dominates. Per-benchmark results and larger canvases are in supp.~G.

%% file: sec/7_conclusion.tex
\section{Conclusion}
\label{sec:conclusion}

Overlapping tiles are how fixed-window ViTs run at deployment resolution, yet most of their duplicated computation is unnecessary. The tile arrangement matters only for attention, and per-tile token states matter at most in the first few layers of MiniMax-H3, MAGI-1 and ViTok-v2. Keeping exactly these layers per tile gives MiniMax-H3 a training-free decoder that is $1.4$--$1.5\times$ faster and equivalent to the released one in PSNR, LPIPS and DISTS, and the same operator carries over to MAGI-1 and ViTok-v2.

%% file: sec/supp_body.tex
Section letters match the references to the supplementary material (supp.~A--L) in the main paper. All $\Delta$ values are relative to the released tiled decoder (V0) unless stated otherwise; CIs are paired bootstrap 95\% intervals over videos or images (10{,}000 resamples, seed 0) unless stated otherwise.

\section{What is exact, operation by operation}
\label{supp:exact}
\Cref{tab:s_exact} lists every operation of a MiniMax-H3 block under \method. The numerical checks use one temporal window of a held-out clip (latent frames 5--11, 44$\times$80 tokens, 28 tiles).
\begin{itemize}
\item \emph{Without overlap} (a 32$\times$48-token crop cut into $2\times3$ tiles), \method at \splitk{0} or \splitk{4} matches per-tile decoding with the released module at a PSNR of 78.61\,dB between the two outputs (max.\ abs.\ difference 0.0054); the released decoder's own per-tile and batched calls differ by 77.86\,dB. In bf16 the two numbers are 57.99 and 57.53\,dB.
\item \emph{With overlap}, when every call is issued with the shapes and memory layouts of the released per-tile calls, the first layer's \ttt{proj\_in}, Norm1, Q/K/V, QK-norm and per-tile SDPA outputs, the register rows, and the blended attention output and block output of every single-tile token (8{,}064 of 24{,}640 tokens) are bit-identical to the released decoder. Overlap tokens differ after blending (relative $\ell_2$ 0.26 for the attention output, 0.099 for the block output).
\item Our batched implementation issues calls with other shapes, for which cuBLAS picks other kernels: every stage then differs by about $3\cdot10^{-4}$ (relative $\ell_2$), the same order as the released decoder's per-tile vs.\ batched calls.
\end{itemize}

Two exact statements underlie these checks. Let $\bar w_{t,i}$ be the released stitching weight pooled to token $i$ in tile $t$, with $\sum_{t\ni i}\bar w_{t,i}=1$.

\emph{Proposition 1 (first layer).} If positions enter the decoder only inside attention and every tile carries its own global tokens, then computing the first block's per-token operations before attention once per token is exact, each tile's first attention output equals the released one, and the first block's output is unchanged for every token covered by a single tile.
\emph{Proof.} The latent tokens and the input projection do not depend on the tile, so $h^{(0)}_{t,i}=h^{(0)}_i$ for all $t\ni i$. Normalization, QKV projections and QK-normalization act on one token at a time and therefore give the same result in every tile. Positions are applied to queries and keys inside each tile's attention with that tile's coordinates, so tile $t$ attends over exactly the queries, keys, values and global tokens of the baseline. For a token in a single tile, the blend has one term with weight 1. \hfill$\square$

\emph{Proposition 2 (sharing QKV and the MLP is full sharing).} From layer $k$ on, let each tile keep its own residual stream $h_{t,i}$, compute QKV once from the weighted average $\bar h_i=\sum_t\bar w_{t,i}h_{t,i}$, apply the out-projection per stream, and compute the MLP once on the weighted average and add it to every stream. Then $\bar h_i$ equals the shared state of \splitk{$k$} at every layer, and so does the output when the streams are averaged before the output head.
\emph{Proof.} At layer $k$, $\bar h_i$ is the merged state of \splitk{$k$}. Suppose $\bar h_i$ equals the shared state at the input of a block. Then QKV, and hence each tile's attention output $a_{t,i}$, are those of \splitk{$k$}. The out-projection $W_O$ is affine and the weights sum to one, so $\sum_t\bar w_{t,i}(h_{t,i}+W_Oa_{t,i}+b_O)=\bar h_i+W_O\tilde a_i+b_O$, with $\tilde a_i$ the blended attention output of \splitk{$k$}. The MLP adds the same update to every stream and hence to $\bar h_i$. Both updates match \splitk{$k$}, which completes the induction. \hfill$\square$

\begin{table*}[t]
\centering\small
\setlength{\tabcolsep}{5pt}
\caption{\textbf{Each operation of a MiniMax-H3 block under \method.}}
\label{tab:s_exact}
\begin{tabular}{>{\raggedright\arraybackslash}p{0.27\textwidth}>{\raggedright\arraybackslash}p{0.33\textwidth}>{\raggedright\arraybackslash}p{0.30\textwidth}}
\toprule
Operation & Layer 0 & Layers $\ge k$ \\
\midrule
\ttt{post\_quant\_conv}, \ttt{proj\_in} & once per token; same input in all tiles, exact & -- \\
Registers / zero token & one set per tile, as released & per tile; attend only to their own tile \\
Norm1, QKV, QK-RMSNorm & once per token; exact & once, on the shared state \\
RoPE & per tile, on tile coordinates, as released & as released \\
Attention (dense, per tile) & released key set, coordinates and registers; exact & keys and values from shared states \\
\textbf{Blend of attention outputs} & \textbf{the only approximation} (overlap tokens) & in every layer \\
Out-projection, residual & exact for single-tile tokens & once per token \\
Norm2, SwiGLU MLP & exact for single-tile tokens & once per token \\
Merge at layer $k$ (\splitk{$k$}) & exact if $k=0$ & weighted average of diverged per-tile states \\
Output head & -- & once per token (no pixel blending) \\
\bottomrule
\end{tabular}
\end{table*}

\section{Full MiniMax-H3 results}
\label{supp:h3}
\paragraph{Data.} DAVIS-30 and 1080p-26 are held-out DAVIS 2017 clips; Dev-8 contains five generated and three real clips used during design; All-38 is DAVIS-30 plus Dev-8. UVG uses all 16 sequences of the UVG dataset~\cite{uvg} (CC BY-NC 3.0), read from the official 8-bit 4:2:0 files from frame 0, subsampled to 30\,fps (120\,fps sources) or 25\,fps (50\,fps sources), area-resized and center-cropped to 704$\times$1280 (121 frames) and 1088$\times$1920 (73 frames; ShakeNDry has 73 frames at both sizes), and encoded with the released MiniMax-H3 encoder. The UVG settings, margins and predictions were written down before any UVG result was computed. The three predictions were: (a) \splitk{4} is equivalent in PSNR at both resolutions; (b) \splitk{6} is equivalent in PSNR at both resolutions and within the strict LPIPS and DISTS margins; (c) \splitk{4} shows excess steps on tile-start lines whose per-video CI excludes 0 at one resolution or more, and \splitk{6} fewer. (a) and (c) held. (b) held for PSNR and DISTS; in LPIPS, the 90\% CIs of \splitk{6} ($[-0.0013, -0.0001]$ at 704p, $[-0.0018, +0.0001]$ at 1080p) extend beyond the margins ($\pm0.00102$, $\pm0.00121$) only toward lower LPIPS; \splitk{6} is therefore non-inferior there, with a lower mean than the released decoder.

\begin{figure*}[t]
\centering
\includegraphics[width=\textwidth]{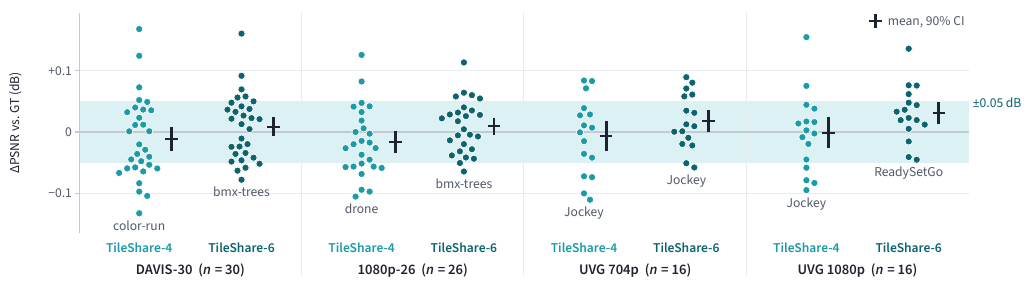}
\caption{\textbf{Per-video $\Delta$PSNR} of \splitk{4} and \splitk{6} relative to the released decoder on DAVIS-30, 1080p-26 and UVG at 704p and 1080p; bars: mean with 90\% CI; shaded: $\pm0.05$\,dB.}
\label{fig:s_pervideo}
\end{figure*}

\paragraph{Equivalence decisions.} With 90\% CIs (margins in parentheses): on DAVIS-30, $\Delta$LPIPS is $+0.0013$ [$+0.0008$, $+0.0017$] for \splitk{4} and $+0.0004$ [$+0.0001$, $+0.0008$] for \splitk{6} ($\pm0.00093$), and $\Delta$DISTS $+0.0007$ [$+0.0005$, $+0.0009$] and $+0.0000$ [$-0.0001$, $+0.0002$] ($\pm0.00049$); on 1080p-26, \splitk{6} has $+0.0002$ [$-0.0001$, $+0.0005$] and $-0.0001$ [$-0.0003$, $-0.0000$] ($\pm0.00093$, $\pm0.00042$). \splitk{4} is within three times the strict margins everywhere. Per-video distributions are in \cref{fig:s_pervideo}.

\begin{table*}[t]
\centering\small
\setlength{\tabcolsep}{4pt}
\caption{\textbf{MiniMax-H3 without training, all evaluation sets.} $\Delta$PSNR (dB). Last two columns: $\Delta$LPIPS and $\Delta$DISTS on All-38 (bold: within the strict margin of 1.16\% of the baseline's mean).}
\label{tab:s_h3}
\resizebox{\textwidth}{!}{\begin{tabular}{lccccccrr}
\toprule
Method & DAVIS-30 & Dev-8 & All-38 [95\% CI] & 1080p-26 & UVG 704p & UVG 1080p & $\Delta$LPIPS & $\Delta$DISTS \\
\midrule
Hard stitching & $-0.199$ & $-0.175$ & $-0.194$ [$-0.209,-0.179$] & -- & -- & -- & $+0.0010$ & -- \\
\splitk{0} & $-0.302$ & $-0.243$ & $-0.290$ [$-0.337,-0.242$] & $-0.353$ & -- & -- & $+0.0037$ & $+0.0030$ \\
\splitk{4} & $-0.011$ & $+0.029$ & $-0.003$ [$-0.026,+0.021$] & $-0.017$ & $-0.006$ & $-0.002$ & $+0.0009$ & $+0.0004$ \\
\splitk{6} & $+0.009$ & $+0.041$ & $+0.015$ [$-0.003,+0.034$] & $+0.009$ & $+0.018$ & $+0.031$ & $\mathbf{+0.0002}$ & $\mathbf{-0.0002}$ \\
FP8 MLP layers only & $-0.008$ & $-0.010$ & $-0.008$ [$-0.014,-0.004$] & -- & -- & -- & $\mathbf{+0.0000}$ & -- \\
\splitk{4} + FP8 & $-0.020$ & $+0.014$ & $-0.012$ [$-0.040,+0.014$] & $-0.026$ & -- & -- & $+0.0009$ & $+0.0004$ \\
\splitk{6} + FP8 & $+0.000$ & $+0.026$ & $+0.006$ [$-0.016,+0.026$] & $+0.000$ & -- & -- & $\mathbf{+0.0002}$ & $\mathbf{-0.0002}$ \\
Spatio-temporal (\cref{supp:st}) & $-0.041$ & $+0.015$ & $-0.029$ [$-0.053,-0.005$] & $-0.044$ & -- & -- & $+0.0010$ & $+0.0004$ \\
\bottomrule
\end{tabular}}
\end{table*}

\begin{table}[t]
\centering\small
\setlength{\tabcolsep}{2.5pt}
\caption{\textbf{Motion strata.} $\Delta$PSNR (dB) by thirds of mean optical-flow magnitude (All-38: boundaries 4.7 / 13.3\,px per frame; 1080p-26: 10.1 / 23.4). $\rho$: Spearman correlation of $\Delta$PSNR with flow magnitude.}
\label{tab:s_motion}
\resizebox{\linewidth}{!}{\begin{tabular}{llcccc}
\toprule
Set & Method & Slow & Medium & Fast & $\rho$ \\
\midrule
All-38 & \splitk{4} & $+0.041$ & $+0.003$ & $-0.051$ [$-0.077,-0.023$] & $-0.50$ \\
All-38 & \splitk{6} & $+0.055$ & $+0.022$ & $-0.030$ [$-0.047,-0.011$] & $-0.60$ \\
All-38 & spatio-temporal & $+0.017$ & $-0.028$ & $-0.076$ [$-0.102,-0.046$] & -- \\
1080p-26 & \splitk{4} & $+0.011$ & $-0.030$ & $-0.032$ [$-0.067,+0.007$] & $-0.45$ \\
1080p-26 & \splitk{6} & $+0.034$ & $+0.002$ & $-0.009$ [$-0.037,+0.022$] & $-0.51$ \\
\bottomrule
\end{tabular}}
\end{table}

\begin{table}[t]
\centering\small
\setlength{\tabcolsep}{2.5pt}
\caption{\textbf{Temporal consistency and zones} (relative to V0). WE: warping error with ground-truth flow (\%), full frame / four-tile corners. Flicker: residual frame difference in four-tile corners ($\times255$). Jump: excess frame difference at temporal-window boundaries ($\times255$).}
\label{tab:s_temporal}
\resizebox{\linewidth}{!}{\begin{tabular}{llccc}
\toprule
Set & Method & WE (\%) & Flicker & Jump \\
\midrule
All-38 & \splitk{4} & $-1.80$ / $-0.68$ & $+0.016$ & $-0.100$ \\
All-38 & \splitk{6} & $-0.83$ / $+0.54$ & $+0.012$ & $-0.035$ \\
All-38 & spatio-temporal & $-0.57$ / $+0.60$ & $+0.043$ & $-0.040$ \\
1080p-26 & \splitk{4} & $-1.26$ / $+1.60$ & $+0.087$ & $-0.102$ \\
1080p-26 & \splitk{6} & $-0.56$ / $+2.23$ & $+0.060$ & $-0.036$ \\
1080p-26 & spatio-temporal & $-0.38$ / $+2.53$ & $+0.112$ & $-0.062$ \\
\bottomrule
\end{tabular}}
\end{table}

\paragraph{Tile-boundary steps.} We split every boundary line of the released tiling into 32-px segments and compute, in every frame, the luminance step across each segment minus the mean step across parallel lines at $\pm2$ and $\pm4$\,px. We count the segments where a variant's excess exceeds V0's by more than 6 gray levels and subtract the segments where V0's excess exceeds the variant's by as much. Control lines are shifted by 24 or 32\,px. For \splitk{4}, the excess lies on tile-start lines (Gen-44 / All-38 / 1080p-26: $+4.80$ / $+2.77$ / $+5.67$\textperthousand), not on tile-end lines ($-0.6$ to $+0.8$) or control lines ($\pm0.4$, except $+1.1$--$1.3$ on one control line at 704p), and 89--98\% of it lies in four-tile corners. There, the released stitching is not a product of two ramps: at the starting row of a lower tile, the upper-left tile's pixel weight drops from 0.92 to 0 and the lower-left tile's rises from 0 to 0.92; V0 itself thus switches tiles abruptly. \splitk{6} reduces the excess to $+2.04$ / $-0.64$ / $+1.34$\textperthousand{} (UVG: $+2.19\to+0.29$ at 704p and $+2.59\to+0.72$ at 1080p). A zero blend weight on each tile's first token only moves the steps one token inward ($-0.07$\,dB).

\paragraph{Spatio-temporal sharing.}
\label{supp:st}
MiniMax-H3's temporal windows overlap by 2 latents. Temporal redundancy is therefore small (each latent is processed 7/5 times in time, about twice in space), and spatial sharing is the main setting; the temporal overlap can be shared as well. Sharing them from the first layer causes a periodic jump at window boundaries; keeping per-window states of the overlap latents up to layer 8 ($k_t=8$) and of the causal padding latent throughout removes the jumps (\cref{tab:s_temporal}). The result decodes $1.73$ / $1.59$ / $1.61\times$ faster than batched tiling ($2.21$ / $2.05$ / $2.07\times$ with FP8) but outside the non-inferiority margin on every set (\cref{tab:s_h3}); it loses more with fast motion ($-0.076$\,dB), raises warping error on Gen-44 by 3.3\% (\splitk{4}: 0.55\%), lowers VBench motion smoothness slightly, and needs 34--41\,GB of peak memory because a whole clip stays resident. Self-distillation with an additional loss on frame differences at window boundaries makes spatio-temporal sharing non-inferior ($-0.012$\,dB, \cref{supp:train}).

\section{Efficiency}
\label{supp:eff}
\paragraph{Protocol.} FP8 runs the MLPs in e4m3 with per-channel weight and per-token activation scales. One RTX PRO 6000 Blackwell Server Edition (96\,GB, sm\_120, 600\,W), driver 580.95.05, CUDA 13.0, PyTorch 2.12.0, cuDNN 9.20, Triton 3.7.0, diffusers 0.40.0. Release precision (fp32 weights, fp16 autocast); \ttt{torch.compile} (default mode, \ttt{dynamic=False}) for every method; first call (with compilation) and two warm-ups, then ten timed runs with synchronization, median reported (interquartile range $\le1.1\%$). Timing covers latents on the GPU to blended pixels on the GPU (post-quant conv, all blocks and tiles, head, pixel stitching, temporal cross-fade). One video per call; no other process was active on the GPU (checked before each method). During the \splitk{5} and \splitk{6} runs, an idle process of another user held 2.4\,GB on the same GPU without measurable activity; the \splitk{4} anchor timed in the same session agrees with the earlier exclusive session within 1\%. Inputs: 704$\times$1280$\times$175, 768$\times$1344$\times$124 and 1088$\times$1920$\times$73 frames (10, 7 and 4 temporal windows; 28, 28 and 60 tiles).

\begin{table}[t]
\centering\small
\setlength{\tabcolsep}{2.5pt}
\caption{\textbf{Decoder time} (s) and speed-up vs.\ batched tiling, RTX PRO 6000. Peak allocated memory (GB) at 704p.}
\label{tab:s_time}
\resizebox{\linewidth}{!}{\begin{tabular}{lcccc}
\toprule
Method & 704p & 768p & 1080p & Mem. \\
\midrule
V0, per tile, eager (released) & 28.31 (0.32) & 27.42 (0.23) & 29.76 (0.26) & 12.6 \\
V0, batched, compiled & 8.92 (1.00) & 6.25 (1.00) & 7.60 (1.00) & 14.9 \\
V0 + FP8 & 6.59 (1.35) & 4.62 (1.35) & 5.63 (1.35) & 16.2 \\
\splitk{0} & 5.28 (1.69) & 4.06 (1.54) & 4.72 (1.61) & 15.7 \\
\splitk{4} & 5.71 (1.56) & 4.34 (1.44) & 5.07 (1.50) & 15.1--16.8 \\
\splitk{6} & 5.92 (1.51) & 4.43 (1.41) & 5.23 (1.45) & 15.1 \\
\splitk{4} + FP8 & 4.37 (2.04) & 3.32 (1.88) & 3.92 (1.94) & 16.4 \\
\splitk{6} + FP8 & 4.54 (1.96) & 3.38 (1.85) & 4.00 (1.90) & 16.4 \\
Spatio-temporal & 5.15 (1.73) & 3.94 (1.59) & 4.71 (1.61) & 41.4 \\
\bottomrule
\end{tabular}}
\end{table}

\begin{table}[t]
\centering\small
\setlength{\tabcolsep}{3pt}
\caption{\textbf{Where the time goes} (704p, GPU kernel time per decode, ms, \ttt{torch.profiler}). Ideal: linear-layer FLOP ratio 0.549 applied to all per-token work, attention unchanged.}
\label{tab:s_prof}
\resizebox{\linewidth}{!}{\begin{tabular}{lrrrr}
\toprule
Component & V0 & \splitk{4} & Saved & Ratio \\
\midrule
MLP GEMMs & 4674 & 2596 & 2079 & 0.56 \\
QKV + out-projection GEMMs & 1596 & 913 & 683 & 0.57 \\
Attention (SDPA) & 851 & 842 & 9 & 0.99 \\
RMSNorm + residual & 626 & 410 & 216 & 0.66 \\
QK-norm + RoPE (+ gather) & 334 & 311 & 23 & 0.93 \\
SwiGLU gate & 597 & 327 & 270 & 0.55 \\
Blend / scatter (new) & 0 & 146 & $-146$ & -- \\
Weight casts, head, merge & 150 & 140 & 9 & -- \\
\midrule
Total & 8828 & 5685 & 3142 & ideal 5297 \\
\bottomrule
\end{tabular}}
\end{table}

\begin{figure*}[t]
\centering
\includegraphics[width=\textwidth]{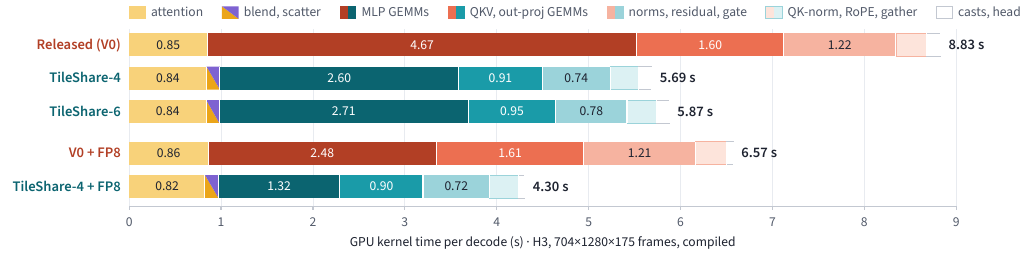}
\caption{\textbf{Where the decoding time goes} (MiniMax-H3, 704$\times$1280$\times$175, compiled, GPU kernel time per decode). Per-token GEMMs and element-wise work shrink in proportion to the token count; attention is unchanged; the blend and scatter of \method are new.}
\label{fig:s_prof}
\end{figure*}

\paragraph{Theory vs.\ measurement.} Linear-layer FLOP ratios are 0.492 (\splitk{0}), 0.549 (\splitk{4}), 0.577 (\splitk{6}) and 0.500 (spatio-temporal) at 704p, giving FLOP bounds of 1.84, 1.69, 1.62 and $1.82\times$; measured speed-ups are 92--95\% of these at all three resolutions (92\% only for \splitk{0} at 704p). GEMM time scales with the token count (\cref{tab:s_prof,fig:s_prof}); the gap of 0.39\,s to the ideal comes from the new blend and scatter (0.15\,s), the gather in the QK-norm/RoPE kernels (0.13\,s), and element-wise kernels that cannot fuse across compiled layer boundaries. Batching two or four temporal windows per call speeds \method up by a further 1--3\% and the baseline not at all. \splitk{6} adds 189\,ms of kernel time over \splitk{4}, mostly in GEMMs.

\paragraph{H100.} On an H100 80GB SXM (Modal, same software and protocol), \splitk{6} reaches $1.50$ / $1.38$ / $1.44\times$ ($1.76$ / $1.64$ / $1.66\times$ with FP8) and \splitk{4} $1.55$ / $1.42$ / $1.49\times$ ($1.81$ / $1.68$ / $1.71\times$); FP8 alone gives only $1.19$--$1.20\times$ there; the share of \method in the combined speed-up is therefore larger than on the RTX PRO 6000. In an earlier session with five timed runs, \splitk{0} reached $1.67$ / $1.51$ / $1.59\times$; \splitk{4} then agreed with the current session within 1.1\%.

\paragraph{End to end.} FastH3 (4 steps, 768$\times$1344$\times$124, one GPU, VAE not offloaded): the denoiser takes 29.2--29.4\,s and decoding 6.45\,s with batched tiling (16\% of the request), 4.54\,s with \splitk{4} and 3.57\,s with \splitk{4} + FP8. Per-request wall-clock time (40.5--41.6\,s) varies by 1.0--4.5\,s between runs because of text encoding with CPU offload, which masks this saving. With one denoiser pass (2 sampling steps, $n=2$), decoding takes 40\% and \splitk{4} shortens the request from 16.1 to 15.0\,s ($1.08\times$; $1.11\times$ with FP8); with two passes, denoiser plus decoder time drops from 23.7 to 18.4\,s with \splitk{4} + FP8 ($1.29\times$). With the released VAE offload, moving 10.4\,GB of fp32 weights between CPU and GPU dominates the decoding stage regardless of the method. 1080p and 243-frame requests exceed the memory of one GPU in the denoiser.

\section{Ablations}
\label{supp:abl}
\begin{table}[t]
\centering\small
\setlength{\tabcolsep}{3pt}
\caption{\textbf{Shared operations, all combinations} (MiniMax-H3, $\Delta$PSNR in dB; $k=4$: All-38; $k=0$: Dev-8, except where All-38 is given in parentheses). q: Norm1 + QKV + QK-norm; o: out-projection; m: MLP. Per-tile residual streams are kept from layer $k$ on.}
\label{tab:s_comp}
\resizebox{\linewidth}{!}{\begin{tabular}{lcccc}
\toprule
Shared & $k=4$ & FLOPs & $k=0$ & FLOPs \\
\midrule
none & $-0.008$ & 1.000 & $-0.008$ & 1.000 \\
q & $-0.070$ & 0.924 & $-2.666$ ($-3.262$) & 0.914 \\
o & $-0.785$ & 0.975 & $-0.243$ & 0.971 \\
m & $-0.016$ & 0.695 & $+0.006$ ($-0.022$) & 0.657 \\
q + o & $-0.036$ & 0.898 & $-0.243$ & 0.886 \\
q + m ($\equiv$ all) & $-0.003$ & 0.619 & $-0.243$ & 0.571 \\
o + m & $+0.001$ & 0.670 & $-0.243$ & 0.628 \\
all (\method) & $-0.003$ & 0.593 & $-0.243$ ($-0.290$) & 0.543 \\
m in layers 0--3, all from 4 & $-0.013$ & 0.555 & -- & -- \\
\bottomrule
\end{tabular}}
\end{table}

\paragraph{Components.} \Cref{tab:s_comp} gives all combinations. By Proposition~2 (\cref{supp:exact}), \{q, m\} equals full sharing at any $k$; at $k=0$, every configuration with a broadcast attention update keeps the per-tile states identical and equals \splitk{0}. A hybrid that shares only the MLP in the first four layers and everything afterwards needs 0.555 of the baseline's FLOPs at $-0.013$ [$-0.039$, $+0.013$]\,dB.

\paragraph{Blending rules} (MiniMax-H3, All-38, $k=4$ / $k=0$): linear ramp $-0.003$ / $-0.290$; Gaussian, $\sigma=0.25$ tile $-0.094$ / $-0.370$; linear at the merge then uniform $-0.195$ / --; Gaussian, $\sigma=0.10$ tile (as in tiled diffusion) $-0.223$ / $-0.779$ (Dev-8); dominant tile $-0.298$ / $-1.214$; uniform $-0.821$ / $-1.494$.

\paragraph{Overlap width} (All-38, block FLOPs relative to V0 at the released overlap, measured speed-up at 704p). V0 with minimum overlap 0 / 16 / 48 / 128\,px: $-0.747$ / $-0.342$ / $-0.187$ / $-0.257$\,dB at 0.536 / 0.643 / 0.857 / 1.607 FLOPs ($1.86$ / $1.55$ / $1.16\times$ / not timed). \splitk{4} at the same overlaps: $-0.812$ / $-0.444$ / $-0.248$ / $-0.464$\,dB, \ie, 0.06--0.21\,dB below each overlap's own V0. The released 64\,px overlap is the only width at which a variant stays within 0.05\,dB.

\paragraph{Temporal only and FP8.} Sharing only the temporal overlap (each spatial tile keeps its own tokens) costs $-0.143$\,dB at $k=4$ and $-0.027$\,dB with $k_t=8$, while saving only 7.5\% of the FLOPs. FP8 MLP layers alone cost $-0.008$\,dB and compose with \method (\cref{tab:s_time}).

\section{Mechanism and the choice of $k$}
\label{supp:mech}
\begin{table}[t]
\centering\small
\setlength{\tabcolsep}{2.5pt}
\caption{\textbf{Per-tile-depth sweep} on MiniMax-H3 (All-38). LPIPS and DISTS as differences; HF: high-frequency energy relative to V0, on float outputs (the main paper's Tab.~4 uses uint8 frames, hence $-4.7$ there for $k=4$); zones: four-tile corners.}
\label{tab:s_ksweep}
\resizebox{\linewidth}{!}{\begin{tabular}{rcccrr}
\toprule
$k$ & $\Delta$PSNR [95\% CI] & $\Delta$LPIPS & $\Delta$DISTS & HF (\%) & Corners \\
\midrule
0 & $-0.290$ [$-0.337,-0.242$] & $+0.0037$ & $+0.0030$ & $-4.5$ & $-0.544$ \\
2 & $-0.223$ [$-0.266,-0.180$] & $+0.0030$ & $+0.0021$ & $-5.3$ & $-0.431$ \\
3 & $-0.113$ [$-0.147,-0.079$] & $+0.0024$ & $+0.0014$ & $-6.4$ & $-0.270$ \\
4 & $-0.003$ [$-0.026,+0.021$] & $+0.0009$ & $+0.0004$ & $-4.8$ & $-0.059$ \\
5 & $+0.013$ [$-0.008,+0.033$] & $+0.0005$ & $+0.0001$ & $-3.9$ & $-0.035$ \\
6 & $+0.015$ [$-0.003,+0.034$] & $+0.0002$ & $-0.0002$ & $-2.6$ & $-0.032$ \\
8 & $-0.048$ [$-0.063,-0.033$] & $-0.0005$ & $-0.0003$ & $+1.3$ & $-0.129$ \\
12 & $-0.047$ [$-0.060,-0.036$] & $-0.0007$ & $-0.0003$ & $+2.2$ & $-0.132$ \\
18 & $-0.046$ [$-0.052,-0.041$] & $-0.0007$ & $-0.0003$ & $+3.4$ & $-0.104$ \\
36 & $-0.008$ [$-0.011,-0.006$] & $+0.0000$ & $+0.0000$ & $+0.2$ & $-0.021$ \\
\bottomrule
\end{tabular}}
\end{table}

\begin{figure}[t]
\centering
\includegraphics[width=\linewidth]{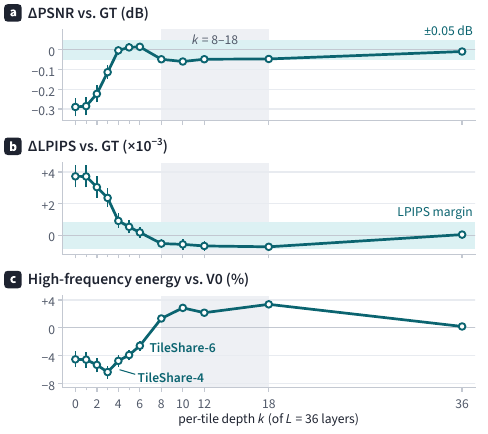}
\caption{\textbf{Distortion and perception along the per-tile-depth sweep} (MiniMax-H3, All-38, 95\% CIs). Between $k=8$ and 18, \splitk{$k$} is about 0.05\,dB below the released decoder in PSNR but sharper than it, with lower LPIPS.}
\label{fig:s_ksweep}
\end{figure}

\paragraph{Per-tile-depth sweep.} \Cref{tab:s_ksweep,fig:s_ksweep} give the MiniMax-H3 sweep. MAGI-1 (10 DAVIS videos): \splitk{0} $-0.203$, \splitk{2} $-0.108$, \splitk{3} $-0.024$ (equivalent), \splitk{4} $-0.016$\,dB; on a second set of 8 videos, $k^\star=3$ as well. Keeping only the last 1, 2 or 4 layers of MiniMax-H3 per tile gives $-0.243$, $-0.243$ and $-0.242$\,dB on Dev-8, identical to \splitk{0}.

\paragraph{Merge-once cost.} Merging the per-tile states once at the input of layer $\ell$ and letting the tiles continue independently changes the output by (PSNR to V0, dB, mean squared error over 30 temporal windows) 42.6, 38.6, 39.6, 42.2, 44.4, 47.8, 52.2, 55.6 and 54.0 in MiniMax-H3 at $\ell=1,2,3,4,8,12,18,24,36$ (layer 0 is a no-op, because the per-tile states of RoPE models are identical there), and by 43.2, 43.1, 46.0, 49.8, 51.5, 55.0, 57.6 and 58.1 in MAGI-1 at $\ell=0,1,2,3,6,12,18,24$ (released setting, 42 windows). An isotropic perturbation of the same norm gives 45.5, 41.0, 42.5, 46.3, 45.7, 44.2, 46.6, 46.2 and 45.9\,dB in MiniMax-H3 at the same layers: the merge changes the output more than the random perturbation up to layer 8 (by 1.3--4.9\,dB) and 3.5--9.7\,dB less from layer 12 on (MAGI-1: 1.3--3.9\,dB more in layers 1--8, 7.0--10.0\,dB less from layer 18 on). The sum of merge costs from layer $k$ on ranks the deviation of \splitk{$k$} from V0 exactly in both decoders (Spearman 1.00; Pearson of the logarithms 0.92 in both); it ranks the loss against ground truth less well in MiniMax-H3 (0.57), whose sweep is not monotone.

\begin{figure}[t]
\centering
\includegraphics[width=\linewidth]{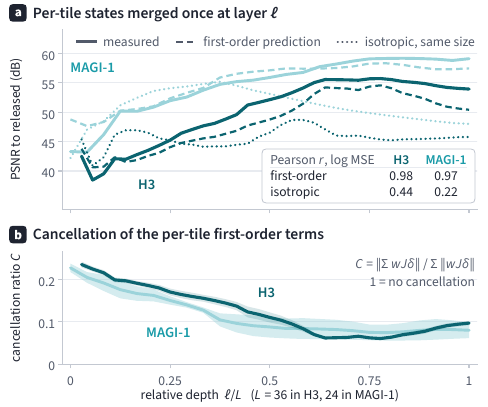}
\caption{\textbf{A first-order account of the merge cost} (MiniMax-H3: 24 temporal windows; MAGI-1: 14 segments). (a) Merging the per-tile states once at layer $\ell$: PSNR of the output to the unmerged output (solid), of the first-order term $\sum_t w_tJ_t\delta_t$ computed by forward-mode differentiation (dashed), and of a random perturbation of the same size on the same windows (dotted). (b) Cancellation ratio $C$ (1: no cancellation; band: $\pm1$ s.d.\ over windows).}
\label{fig:s_taylor}
\end{figure}

\paragraph{A first-order account of the merge cost.} Let $G_t$ be the rest of tile $t$'s network from layer $\ell$ on, $J_t$ its Jacobian at the unmerged states, and $\delta_t=\bar h-h_t$ the displacement that the merge applies to an overlap token's state in tile $t$. With the released pixel weights $w_t$, the merge changes the output by
\[\Delta y=\textstyle\sum_t w_t\big[G_t(\tilde H_t)-G_t(H_t)\big]=\sum_t w_t J_t\delta_t+O(\lVert\delta\rVert^2).\]
Because $\sum_t w_t\delta_t=0$, the first-order term vanishes when all tiles respond to the token in the same way and otherwise equals $\sum_t w_t(J_t-\bar J)\delta_t$: it measures how differently the tiles respond, along the directions in which their states move. A random perturbation of the same size has no such cancellation, which is why it does not predict the merge cost. We compute the first-order term exactly by forward-mode differentiation in fp32 (MiniMax-H3: 24 of the 30 windows; MAGI-1: 14 of the 42 segments). Across layers, it tracks the measured merge cost (\cref{fig:s_taylor}a): the Pearson and Spearman correlations of the log mean squared errors are 0.98 and 1.00 in MiniMax-H3 and 0.97 and 0.95 in MAGI-1, against 0.44 and 0.27, and 0.22 and $-0.28$, for the random perturbation on the same windows. The cancellation ratio $C=\lVert\sum_t w_tJ_t\delta_t\rVert/\sum_t \lVert w_tJ_t\delta_t\rVert$, with the weights applied per pixel, falls from 0.22 and 0.20 in the first four layers to 0.08 in the last third of both decoders (\cref{fig:s_taylor}b): deep layers respond alike in every tile, and merging there is nearly free to first order. The first-order term accounts for most of the output change in deep layers (at least 70\% of it for $\ell=22$--31 in MiniMax-H3 and $\ell=15$--24 in MAGI-1). In the first layers, the merge displacement lies beyond the linear regime (central differences at half its size differ from the derivative by 25\% in MiniMax-H3 and 57\% in MAGI-1), and the first-order term underestimates the cost there, by up to 5.5\,dB in MAGI-1.

\paragraph{Attention statistics.} In MAGI-1, the attention mass that overlap tokens place on keys shared by both tiles falls from 0.65 (layer 2) to 0.31 (last layer) and tracks the merge cost (Spearman 0.92 across layers), as do attention entropy ($-0.90$) and distance ($-0.86$). In MiniMax-H3 these statistics fluctuate without trend. Neither decoder attends preferentially to tile borders.

\paragraph{Predicting $k$.} One sweep over a few depths on a few development clips measures $k$ directly; the predictors below do not replace it. We measured, on all 14 configurations of the main paper, the divergence of the per-tile states at the input ($\delta_0$, zero for RoPE models) and tile agreement on the baseline. With two thresholds fitted to the data, they place configurations into $k^\star=0$ / small / large with 12 of 14 correct in sample and 10 of 14 in leave-one-out; the errors are PE-Spatial and SBB ($k^\star=0$ despite low agreement) and MiniMax-H3 and RAE-MAE, which flip near a threshold. A rule without ground truth that picks the smallest $k$ whose deviation from V0 is below that caused by shifting the input by one token (fixed in advance) picks the right grid point in 10 of 14 configurations, but six of these hits are models whose deviation never meets the threshold, for which the rule returns $L$; among the seven models with $k^\star\le L/6$ it is right for three (MiniMax-H3, MAGI-1, RAE-MAE) and too deep for ViTok-v2 350M and 5B, PE-Spatial and SBB. A variant calibrated to FP8's deviation is right in 11 of 12 applicable configurations; it was written down after a few samples and does not apply to MiniMax-H3 and MAGI-1. A rule that requires halving the deviation of \splitk{0} is right in 7 of 14.

\section{Applicability details}
\label{supp:scope}
\paragraph{Tilings.} ViTok-v2 and RAE are released for single-window coding. On 40 DIV8K images~\cite{div8k} cropped to 1K--4K, we tile their encoder and decoder together with tiles of the training size and 25\% overlap blended with linear ramps. At 4K, whole-image coding with ViTok-v2 350M is 2.0\,dB worse than overlapping tiled coding. Encoders use linear segmentation heads on ADE20K~\cite{ade20k} at short side 512 with 512\,px windows and stride 341 (the official head for DINOv2, short side 518), and the CLIP methods use their released settings (SCLIP and NACLIP: 224\,px windows, stride 112, short side 336).

\paragraph{Criteria for $k^\star$.} Without training, $k^\star$ is the smallest per-tile depth that is non-inferior for video decoders (95\% CI above $-0.05$\,dB) and encoders (above $-0.3$ mIoU); for image decoders we use the weaker criterion of a mean loss below 0.05\,dB or a CI containing 0.

\paragraph{Resolution.} At short side 896 (overlap $R\approx2.8$), $k^\star$ grows for encoders: DINOv3-B 11 of 12, DINOv3-L 16 of 24, PE-Spatial 6 of 12, SBB 6 of 12; whole-image inference is then better than sliding windows for DINOv3 ($+1.6$ / $+1.7$ mIoU) and PE-Spatial ($+2.5$) but worse for SBB ($-6.1$).

\paragraph{Positions vs.\ context in the RAE decoders.} In the RAE decoders, context rather than position drives the deviation: slicing one global position embedding instead of per-tile ones makes every token's per-tile states identical at the input, yet \splitk{0} still deviates from V0 at 26.7\,dB (24.2\,dB with per-tile embeddings; RAE-MAE, without position embeddings: 46.1\,dB); the sliced baseline itself loses $3.07$\,dB (DINOv2-B) and $4.54$\,dB (SigLIP2).

\paragraph{Sharing only the MLP} from layer 0 (main paper, Sec.~4.1) does not replace a large per-tile depth (95\% CIs, same data and criteria as Tab.~5 of the main paper). In the RAE decoders its $\Delta$PSNR is $-0.29$ [$-0.40,-0.19$] (SigLIP2-B), $-0.42$ [$-0.57,-0.29$] (DINOv2-B) and $-0.44$ [$-0.57,-0.32$]\,dB (DINOv2-L), although it removes 62--82\% of the loss of \splitk{0}; for DINOv2-B and CLIP with SCLIP, $\Delta$mIoU is $-0.87$ [$-1.57,-0.43$] and $-2.62$ [$-2.98,-2.18$]. For DINOv3-B and -L at the standard resolution it meets the criterion ($-0.08$ and $-0.10$ mIoU) at 0.85 of the block FLOPs, against 0.92 for \splitk{$k^\star$}. In MiniMax-H3 it decodes $1.40$ / $1.32$ / $1.36\times$ faster than batched tiling at 704p / 768p / 1080p, against $1.50$ / $1.40$ / $1.45\times$ for \splitk{6} in the same session, with 1.3--2.5\,GB less peak memory.

\section{Training details}
\label{supp:train}
\paragraph{LoRA self-distillation (MiniMax-H3).} Rank-32 LoRA ($\alpha=32$; A Kaiming-initialized, B zero) on all linear layers (51.9M parameters), AdamW ($\beta=0.9/0.99$, no weight decay, learning rate $2\cdot10^{-4}$, 50 warm-up steps, cosine decay to 5\%, gradient clipping at 1.0), 2500 steps, batch 1, random latent crops of $3\times3$ to $4\times4$ tiles, loss L1 + $0.1\cdot$LPIPS on four random frames, teacher = frozen V0 with released tiling; 60 DAVIS training sequences and 30 FastH3 latents, disjoint from all test videos and prompts. Merged, the student runs at the speed of \splitk{0} ($2.00$--$2.20\times$ with FP8, timed in an earlier session). A second seed gives $+0.008$\,dB with per-video correlation 0.99; on 1080p-26 the student gives $-0.004$\,dB [$-0.042$, $+0.027$] and halves the corner loss of \splitk{4}. Running the adapted weights as an ordinary tiled decoder reduces cross-tile divergence by only 7--10\% and costs 0.03\,dB; with an unseen denser tiling (128\,px overlap) the student recovers 70\% of the \splitk{0} gap. For spatio-temporal sharing, the same recipe removes the quality gap but not the window-boundary jumps; adding a loss on frame differences at window boundaries gives $-0.012$\,dB on All-38 with jumps at the baseline level ($1.93$ / $1.75\times$; $2.44$ / $2.23\times$ with FP8). With less overlap, the same recipe gives $-0.150$\,dB [$-0.182,-0.115$] at 16\,px (worst zone $-0.33$, $1.55\times$) and $-0.387$\,dB [$-0.441,-0.328$] at 0\,px (band around its tile borders $-1.21$, $1.86\times$). On MAGI-1 (16-frame chunks, whole-frame output head; 10 DAVIS videos), the recipe closes 61\% of the \splitk{0} gap ($-0.203\to-0.079$\,dB, 12.6M parameters), and the coordinate field 75\% ($-0.050$\,dB, 1.2M parameters, $1.11\times$).

\paragraph{Ground-truth fine-tuning.} We distill from the released decoder rather than fine-tune on ground truth: LoRA fine-tuning of V0 and \splitk{0} against ground truth (L1 + $0.2\cdot$LPIPS) raises the PSNR of both by blurring (high-frequency ratio 0.80 $\to$ 0.58), with the shared variant 0.145\,dB behind.

\paragraph{Coordinate field.} Inputs are Fourier features of each token's global coordinates (log-spaced periods, the released tile-grid periods, ramps to the frame border, normalized coordinates and their harmonics; for MiniMax-H3 also coverage indicators and blend weights) and, for video, the latent frame index within its window. A two-layer MLP of width 256 produces $K$ basis fields ($K=2$ for MiniMax-H3, 4 for MAGI-1); each layer adds $\sum_K g_{\ell,K}\odot b_K$ to the residual stream of every unique token through zero-initialized gates $g$. The field is trained with the self-distillation loss above (MiniMax-H3: 2500 steps, learning rate chosen from three values on training crops only; MAGI-1: 3000 steps) and precomputed per latent shape at inference, which adds one addition per layer (2\% decoder time on MAGI-1). In MiniMax-H3, the injected field is 5--10\% of the residual norm in the first five layers and decays with depth; it is 1.5$\times$ larger in four-tile corners than in single-tile regions.

\paragraph{CLIP.} The field (0.64M parameters) is trained for 12{,}000 steps with SCLIP as the teacher (per-window features of the frozen model, count-averaged), on ADE20K and VOC training images at short sides 336--672, and selected only by cosine similarity on held-out training images. For ClearCLIP (448\,px windows, stride 224), the field's absolute coordinates are divided by 2, a factor chosen on held-out training images. \Cref{tab:s_clip} gives every benchmark. Backbone time (compiled, batch 16) falls by $1.71$, $2.15$ and $2.02\times$ at 336, 448 and 560 (FLOP bounds 1.91, 2.56, 2.63; whole-image inference 1.89, 2.45, 2.05$\times$). End to end on one ADE20K image (batch 1), the released per-window loop takes 28.2\,ms, batched windows 11.9\,ms, \splitk{0} with the field 10.1\,ms and whole-image inference 9.2\,ms; the backbone is only 24--27\% of the total. At larger canvases the field falls behind sliding windows on ADE20K ($-0.38$ and $-0.71$ mIoU at short sides 448 and 560 with SCLIP; VOC21: losses of at most 0.3).

\begin{table}[t]
\centering\small
\setlength{\tabcolsep}{2pt}
\caption{\textbf{CLIP open-vocabulary segmentation} ($\Delta$mIoU vs.\ sliding windows). W: whole image. $k_0$: \splitk{0} without the field. F: \splitk{0} with the coordinate field trained on SCLIP.}
\label{tab:s_clip}
\resizebox{\linewidth}{!}{\begin{tabular}{l|cccc|cccc|cccc}
\toprule
& \multicolumn{4}{c|}{SCLIP} & \multicolumn{4}{c|}{NACLIP} & \multicolumn{4}{c}{ClearCLIP} \\
Benchmark & V0 & W & $k_0$ & F & V0 & W & $k_0$ & F & V0 & W & $k_0$ & F \\
\midrule
VOC21 & 59.6 & $-1.0$ & $-2.2$ & $+0.2$ & 57.6 & $+1.2$ & $-2.2$ & $+0.2$ & 51.8 & $-0.9$ & $-1.4$ & $+1.0$ \\
Context60 & 32.6 & $-2.2$ & $-1.4$ & $+0.0$ & 33.4 & $-0.4$ & $-1.9$ & $-0.0$ & 33.8 & $-1.1$ & $-1.1$ & $+0.1$ \\
COCO-Obj. & 33.5 & $-1.7$ & $-2.2$ & $+0.1$ & 33.2 & $-0.3$ & $-2.4$ & $-0.2$ & 33.0 & $-2.1$ & $-1.7$ & $-0.6$ \\
VOC20 & 81.5 & $+0.6$ & $-2.2$ & $+0.8$ & 77.1 & $+2.6$ & $-2.5$ & $+0.9$ & 80.9 & $-1.5$ & $-1.7$ & $-0.1$ \\
Context59 & 34.1 & $-2.4$ & $-1.8$ & $+0.0$ & 35.2 & $-0.3$ & $-2.1$ & $-0.0$ & 35.8 & $-1.1$ & $-1.2$ & $+0.2$ \\
COCO-Stuff & 22.8 & $-3.3$ & $-3.4$ & $-0.1$ & 23.3 & $-0.7$ & $-1.8$ & $-0.2$ & 23.9 & $-1.4$ & $-1.2$ & $-0.1$ \\
ADE20K & 16.5 & $-2.4$ & $-1.0$ & $+0.1$ & 17.4 & $-0.6$ & $-1.0$ & $+0.0$ & 16.7 & $-0.9$ & $-0.7$ & $-0.1$ \\
\midrule
Mean & 40.1 & $-1.8$ & $-2.0$ & $+0.2$ & 39.6 & $+0.2$ & $-2.0$ & $+0.1$ & 39.4 & $-1.3$ & $-1.3$ & $+0.1$ \\
\bottomrule
\end{tabular}}
\end{table}

\section{Downstream models, generation metrics and rFVD}
\label{supp:down}
\paragraph{Downstream consistency} (All-38, every fourth frame). We apply Depth Anything V2 Large~\cite{depthanythingv2} (AbsRel after per-frame scale-and-shift alignment), SegFormer-B5 on ADE20K~\cite{segformer} (pixel disagreement) and RAFT-large~\cite{raft} (EPE on consecutive frames) to each variant's frames, V0's frames and ground-truth frames. The distance between a variant's and V0's predictions is 0.0151 / 0.0234 / 0.382 for \splitk{4} and 0.0130 / 0.0197 / 0.346 for \splitk{6}, against 0.0524 / 0.0726 / 1.020 between V0's and ground truth's. The distance to ground-truth predictions changes by $+0.0012$ / $+0.0010$ (\splitk{4}) and $+0.0004$ / $+0.0007$ (\splitk{6}) in AbsRel / disagreement, CIs containing 0; on 1080p-26 the disagreement rises by $+0.0023$ (\splitk{4}) and $+0.0014$ (\splitk{6}), CIs excluding 0.

\paragraph{Generated videos} (Gen-44, VBench prompts, FastH3, 768$\times$1344$\times$124). PSNR to V0 on uint8 frames: \splitk{4} 41.5\,dB (minimum 37.2), \splitk{6} 42.7\,dB (minimum 38.1). VBench~\cite{vbench} differences (\splitk{4} / \splitk{6}): imaging quality $-0.004$ / $+0.032$ (MUSIQ, 0--100), subject consistency $+0.0002$ / $+0.0002$, temporal flickering $+0.0001$ / $+0.0000$, motion smoothness $-0.00003$ / $-0.00006$.

\paragraph{rFVD} (I3D features, four 16-frame clips and three square crops per video). V0 has 9.64 to ground truth on All-38; \splitk{4} 10.48 (basic bootstrap CI of the difference [$-0.02$, $+1.02$]); the distance between \splitk{4} and V0 is 1.26. With 456 clips, these numbers support only relative statements.

\section{Naive deduplication}
\label{supp:naive}
\begin{table}[t]
\centering\small
\setlength{\tabcolsep}{4pt}
\caption{\textbf{Naive deduplication of MiniMax-H3.} $\Delta$PSNR (dB); tile sizes and whole frame on 18 clips, overlap widths on All-38, sliding window on Dev-8. Speed-up vs.\ batched and compiled tiling at 704p / 768p.}
\label{tab:s_naive}
\resizebox{\linewidth}{!}{\begin{tabular}{lcc}
\toprule
Change & $\Delta$PSNR & Speed-up \\
\midrule
Whole-frame decoding & $-11.93$ & $0.99$ / $0.81$ \\
Larger tiles (320 / 384 / 512\,px) & $-2.41$ / $-6.86$ / $-9.03$ & $0.93$--$1.41$ \\
Less overlap (16 / 0\,px) & $-0.342$ / $-0.747$ & $1.55$ / $1.86$ (704p) \\
Temporal re-chunking & $-3$ to $-9$ & -- \\
Best single-pass sliding window & $-0.67$ & $1.01$ / $0.92$ \\
\bottomrule
\end{tabular}}
\end{table}
\Cref{tab:s_naive} summarizes the naive alternatives. Generic token reduction saves little while keeping PSNR to V0 above 40\,dB on three videos: merging $2\times2$ token groups after block 27 saves at most 4.7\% of the token-block computation, freezing tokens (early exit) 0.28\%, and skipping one block 2.8\%. The best single pass uses global coordinates, a $\pm11$-token attention window, entropy scaling~\cite{cca} and registers read from overlapping tiles; no local-attention kernel makes it faster than batched tiling on our GPU. The seams of such single passes come from the registers. With a sliding window of radius 7, attention has no block borders, yet registers computed per $16\times16$-token query block raise the gradient excess at the block borders to 4.00 (released decoder $-0.19$, $\times255$) and lower PSNR within 2\,px of them to 21.9\,dB; defining the registers over the released tiles and blending them removes the seams (0.09). Narrower overlaps fail for another reason: a single tile's prediction degrades toward its border. In its outermost 8\,px, the squared error is 1.64 times that of the blended output (1.01--1.03 within 64\,px of the tile center), and less overlap gives these predictions more weight.

\section{Reproducibility}
\label{supp:repro}
\emph{Models.} MiniMax-H3 VAE (\ttt{MiniMaxAI/MiniMax-H3}, \ttt{vae/}; SHA-256 of the three weight shards verified against the Hugging Face LFS objects), the FastH3 4-step pipeline (FastVideo 0.2.1, commit \ttt{37d06a8}), the MAGI-1 VAE, ViTok-v2 350M and 5B, the RAE decoders (MAE-B, SigLIP2-B, DINOv2-B/L), DINOv3-B/L (\ttt{facebookresearch/dinov3} at \ttt{6876159a}), DINOv2-B/14 with its official ADE20K linear head (\ttt{facebookresearch/dinov2} at \ttt{7764ea0f}), PE-Spatial-B/16, SBB ViT-betwixt/16 (\ttt{timm}), and CLIP ViT-B/16 with SCLIP (\ttt{36083602}), NACLIP (\ttt{7f2bb0ab}) and ClearCLIP (\ttt{ad68a404}).
\emph{Software.} PyTorch 2.12.0 with CUDA 13.0, cuDNN 9.20, Triton 3.7.0, diffusers 0.40.0, Python 3.12; driver 580.95.05.
\emph{Data.} DAVIS 2017, UVG (CC BY-NC 3.0, official 8-bit files), DIV8K, Commons-40 (supp.~K), ADE20K, PASCAL VOC and Context, COCO-Stuff; the clip lists and preprocessing are released with the code.
We will release the implementation of \method (gather/blend operator and \splitk{$k$} for MiniMax-H3, MAGI-1 and ViTok-v2), the evaluation scripts and the per-video results.

\section{MAGI-1 and ViTok-v2}
\label{supp:others}
\paragraph{MAGI-1.} The released pipeline decodes in temporal chunks of 12 frames (half its frame rate of 24) and with a per-tile output head: each tile unpatches its own tokens and applies the final Conv3d, zero-padded at the tile border, before the pixels are blended. We use exactly this head, so that at $k=L$ \method is bit-identical to batched tiled decoding and the approximate steps are those of the main paper. The MAGI-1 rows of Sec.~6 (supp.~G) use the function's default of 16 frames and apply the head once to the untiled feature map; that head costs 0.004\,dB on DAVIS-10 and 0.017\,dB on UVG relative to the released one. We chose $k=3$ in that setting on DAVIS-10 and eight other clips and froze it, with the released head, before evaluating UVG (preregistered). Re-sweeping $k$ in the released setting afterwards leaves $k^\star=3$ and the shape of the curve unchanged (Pearson 0.998 across the ten depths): $\Delta$PSNR [90\% CI] on DAVIS-10 is $-0.180$ [$-0.212,-0.150$] at $k=0$, $-0.104$ [$-0.123,-0.087$] at $k=2$, $-0.029$ [$-0.035,-0.023$] at $k=3$, $-0.023$ [$-0.028,-0.018$] at $k=4$ and $-0.010$ [$-0.013,-0.007$] at $k=6$; Fig.~3 of the main paper uses this sweep. With the released setting, \splitk{3} gives on DAVIS-10 $\Delta$PSNR $-0.029$ [$-0.035,-0.023$], $\Delta$LPIPS $+0.0007$ (margin $\pm0.0018$) and $\Delta$DISTS $+0.0001$ ($\pm0.0011$); on UVG-16 at 704$\times$1280, $-0.027$ [$-0.032,-0.023$] (worst sequence $-0.050$), $\Delta$LPIPS $+0.0014$ [$+0.0008,+0.0020$] (margin $\pm0.0015$) and $\Delta$DISTS $-0.0004$ [$-0.0006,-0.0001$] ($\pm0.0013$). In the 16-frame setting, UVG gives $-0.011$\,dB with $\Delta$LPIPS $+0.0034$ and $\Delta$DISTS $+0.0018$. Timing follows the main protocol: at 704$\times$1280$\times$176 and 720$\times$720$\times$176, tiled decoding takes 3.34 and 1.89\,s and \splitk{3} 2.96 and 1.66\,s ($1.13$ and $1.14\times$; cost bound $1.32$ and $1.30\times$), with peak memory 24.1 and 21.3\,GB against 14.8 and 16.1\,GB, because the token states of the whole clip stay resident. On the tile-start row inside four-tile corners, \method leaves a faint one-pixel line: the mean luminance step across it is 4.1--5.2 gray levels against 2.5--4.1 for tiled decoding and 3.0--4.2 in the ground truth (one corner crop per video); elsewhere the steps are equal.

\paragraph{ViTok-v2 pipeline.} ViTok-v2 (code at \ttt{a90f23bd}) is released for whole-image coding. We tile its encoder and decoder with the same tiles of $16\times16$ tokens (256\,px) and 25\% overlap; encoder latents of overlap tokens are averaged with the token weights, and every token thus enters the decoder with one state. The baseline decodes all tiles in one batched call and blends pixels with linear ramps; the token weights of \method are the pixel weights averaged over each token's $16\times16$ footprint. All runs use bf16; the authors' own evaluation additionally quantizes to FP8, which we do not. On a single 256\,px tile, \method is bit-identical to the released coder.

\paragraph{Data and preregistration.} DIV8K-40 are the 40 images of the audit (\cref{supp:scope}). Commons-40 was selected by a rule fixed before any download: members of the Wikimedia Commons category \emph{Quality images} added since 2025-01-01, scanned in batches of 50 in the order they were added and by title within a batch (1278 scanned); kept are JPEGs with a short side of at least 4096\,px, an aspect ratio of at most 2, a Creative Commons or public-domain license, and at most one image per uploader; the first 40 are used. Licenses: CC BY-SA 4.0 (31), CC BY 4.0 (4), CC0 (2), public domain (2), CC BY-SA 3.0 (1); files were checked against the API's size and SHA-1. The images are not redistributed; the list with authors and licenses is released with the code. Before the $k$ sweep we wrote down the margins, the rule for choosing $k$ (the smallest $k$ equivalent in all three metrics, otherwise the smallest equivalent in PSNR at both resolutions), the confirmation set, five predictions and the rule for the qualitative figure; $k$ was then written mechanically to a separate file before Commons-40 was evaluated, and the 5B model uses this $k$ without a sweep of its own. The timing runs and a smoke test on one development image, which already showed a lower LPIPS for \method, preceded the preregistration; earlier probes on DIV8K-40 had also suggested $k=0$. The test of the frozen setting is therefore the confirmation set. All five predictions held. On Commons-40, \splitk{0} is equivalent in PSNR at both resolutions and better than tiled decoding in LPIPS and DISTS by more than the margins; tiled coding is 2.03 and 1.25\,dB better than whole-image coding at 4096 and 2048 (predicted: 0.8--3.0 and 0.2--1.5\,dB), and whole-image coding shows a token grid; whole and tiled coding of the 5B model differ by less than 0.3\,dB; and no $k$ is equivalent in all three metrics, because every tested $k$ improves LPIPS and DISTS beyond the margins.

\begin{table}[t]
\centering\small
\setlength{\tabcolsep}{2pt}
\caption{\textbf{ViTok-v2 in detail.} Difference to tiled coding [90\% CI], 40 images each; equivalence marks for PSNR / LPIPS / DISTS as in the main paper (LPIPS and DISTS margins: 0.00035--0.00046 and 0.00023--0.00040). Whole: the released whole-image coding. Hard: each pixel from its dominant tile. +enc: \method also in the encoder.}
\label{tab:s_vitok}
\resizebox{\linewidth}{!}{\begin{tabular}{llccc}
\toprule
Set, size & Variant & $\Delta$PSNR & $\Delta$LPIPS\,/\,$\Delta$DISTS ($10^{-3}$) & P\,/\,L\,/\,D \\
\midrule
DIV8K, 2048 & Whole & $-0.945$ [$-1.292,-0.653$] & $-0.0$\,/\,$-1.3$ & $\times$\,/\,$\times$\,/\,$\uparrow$ \\
 & Hard & $+0.009$ [$-0.001,+0.020$] & $-0.2$\,/\,$-0.3$ & \checkmark\,/\,\checkmark\,/\,$\uparrow$ \\
 & \splitk{0} & $+0.023$ [$+0.014,+0.031$] & $-1.0$\,/\,$-1.6$ & \checkmark\,/\,$\uparrow$\,/\,$\uparrow$ \\
 & \splitk{2} & $+0.023$ [$+0.015,+0.031$] & $-1.0$\,/\,$-1.6$ & \checkmark\,/\,$\uparrow$\,/\,$\uparrow$ \\
 & \splitk{4} & $+0.020$ [$+0.012,+0.028$] & $-1.1$\,/\,$-1.7$ & \checkmark\,/\,$\uparrow$\,/\,$\uparrow$ \\
 & \splitk{8} & $+0.051$ [$+0.042,+0.061$] & $-0.5$\,/\,$-1.1$ & $\uparrow$\,/\,$\uparrow$\,/\,$\uparrow$ \\
 & \splitk{0}+enc & $+0.067$ [$+0.058,+0.075$] & $-1.2$\,/\,$-1.7$ & $\uparrow$\,/\,$\uparrow$\,/\,$\uparrow$ \\
DIV8K, 4096 & Whole & $-1.999$ [$-2.548,-1.540$] & $+4.4$\,/\,$+1.6$ & $\times$\,/\,$\times$\,/\,$\times$ \\
 & Hard & $+0.012$ [$+0.004,+0.021$] & $-0.1$\,/\,$-0.2$ & \checkmark\,/\,\checkmark\,/\,\checkmark \\
 & \splitk{0} & $+0.024$ [$+0.016,+0.033$] & $-1.0$\,/\,$-1.6$ & \checkmark\,/\,$\uparrow$\,/\,$\uparrow$ \\
 & \splitk{2} & $+0.025$ [$+0.016,+0.034$] & $-0.9$\,/\,$-1.6$ & \checkmark\,/\,$\uparrow$\,/\,$\uparrow$ \\
 & \splitk{4} & $+0.021$ [$+0.012,+0.030$] & $-1.0$\,/\,$-1.7$ & \checkmark\,/\,$\uparrow$\,/\,$\uparrow$ \\
 & \splitk{8} & $+0.050$ [$+0.041,+0.060$] & $-0.4$\,/\,$-1.0$ & $\uparrow$\,/\,$\uparrow$\,/\,$\uparrow$ \\
 & \splitk{0}+enc & $+0.064$ [$+0.055,+0.073$] & $-1.1$\,/\,$-1.6$ & $\uparrow$\,/\,$\uparrow$\,/\,$\uparrow$ \\
Commons, 2048 & Whole & $-1.254$ [$-2.195,-0.596$] & $+12.3$\,/\,$+5.3$ & $\times$\,/\,$\times$\,/\,$\times$ \\
 & Hard & $+0.013$ [$+0.004,+0.024$] & $-0.2$\,/\,$-0.3$ & \checkmark\,/\,\checkmark\,/\,$\uparrow$ \\
 & \splitk{0} & $+0.026$ [$+0.017,+0.037$] & $-1.1$\,/\,$-1.9$ & \checkmark\,/\,$\uparrow$\,/\,$\uparrow$ \\
 & \splitk{0}+enc & $+0.064$ [$+0.052,+0.078$] & $-1.4$\,/\,$-2.0$ & $\uparrow$\,/\,$\uparrow$\,/\,$\uparrow$ \\
Commons, 4096 & Whole & $-2.033$ [$-2.730,-1.460$] & $+13.1$\,/\,$+5.2$ & $\times$\,/\,$\times$\,/\,$\times$ \\
 & Hard & $+0.009$ [$+0.002,+0.017$] & $-0.1$\,/\,$-0.2$ & \checkmark\,/\,\checkmark\,/\,\checkmark \\
 & \splitk{0} & $+0.024$ [$+0.014,+0.033$] & $-1.1$\,/\,$-1.8$ & \checkmark\,/\,$\uparrow$\,/\,$\uparrow$ \\
 & \splitk{0}+enc & $+0.060$ [$+0.049,+0.072$] & $-1.2$\,/\,$-1.8$ & $\uparrow$\,/\,$\uparrow$\,/\,$\uparrow$ \\
\midrule
5B: DIV8K, 4096 & Whole & $+0.100$ [$+0.030,+0.177$] & $-4.4$\,/\,$-3.5$ & $\uparrow$\,/\,$\uparrow$\,/\,$\uparrow$ \\
 & \splitk{0} & $+0.019$ [$+0.013,+0.025$] & $-1.6$\,/\,$-2.0$ & \checkmark\,/\,$\uparrow$\,/\,$\uparrow$ \\
5B: Commons, 4096 & Whole & $+0.175$ [$+0.087,+0.271$] & $-4.5$\,/\,$-5.2$ & $\uparrow$\,/\,$\uparrow$\,/\,$\uparrow$ \\
 & \splitk{0} & $+0.013$ [$+0.006,+0.020$] & $-1.9$\,/\,$-2.7$ & \checkmark\,/\,$\uparrow$\,/\,$\uparrow$ \\
\bottomrule
\end{tabular}}
\end{table}

\begin{table}[t]
\centering\small
\setlength{\tabcolsep}{3pt}
\caption{\textbf{ViTok-v2 timing} (ms, median of ten compiled runs, bf16, one image, RTX PRO 6000 with no other process). Decoder: latents to blended pixels; pipeline: encoder and decoder. In parentheses: \method also in the encoder.}
\label{tab:s_vitok_time}
\resizebox{\linewidth}{!}{\begin{tabular}{llccccc}
\toprule
Model, size & Scope & Tiled & Whole & \method & vs.\ tiled & vs.\ whole \\
\midrule
350M, 2048 & decoder & 76.9 & 110.2 & 43.5 & $1.77\times$ & $2.54\times$ \\
 & pipeline & 89.9 & 128.3 & 55.7 (50.5) & $1.61\times$ ($1.78\times$) & $2.30\times$ ($2.54\times$) \\
350M, 4096 & decoder & 271.9 & 1310.6 & 178.4 & $1.52\times$ & $7.35\times$ \\
 & pipeline & 322.8 & 1553.9 & 229.2 (216.5) & $1.41\times$ ($1.49\times$) & $6.78\times$ ($7.18\times$) \\
5B, 4096 & decoder & 3036.5 & 7828.7 & 1885.2 & $1.61\times$ & $4.15\times$ \\
 & pipeline & 3401.9 & 8636.5 & 2212.1 (2089.5) & $1.54\times$ ($1.63\times$) & $3.90\times$ ($4.13\times$) \\
\bottomrule
\end{tabular}}
\end{table}

\paragraph{Non-overlapping tiles.} Coding each 256\,px tile on its own without overlap is the fastest tiled pipeline. Its decoder runs $1.12\times$ ($2048^2$) and $1.15\times$ ($4096^2$) faster than \method, and its PSNR is 0.04--0.06\,dB higher. It leaves a one-pixel discontinuity at every tile border. On tile-border lines, the gradient excess over other token-grid lines is $+1.0$ to $+1.4$ ($\times255$), against none for tiled decoding and \method, and PSNR within 2\,px of the borders is 0.6--0.9\,dB below tiled decoding; its DISTS is also higher than that of \method (\eg, 0.0275 vs.\ 0.0263 on Commons-40 at $4096^2$). \method recovers 80--88\% of the speed difference between overlapping and non-overlapping tiling without these discontinuities. Non-overlapping 512\,px tiles fail on one Commons image, as whole-image coding does.

\begin{figure*}[t]
\centering
\includegraphics{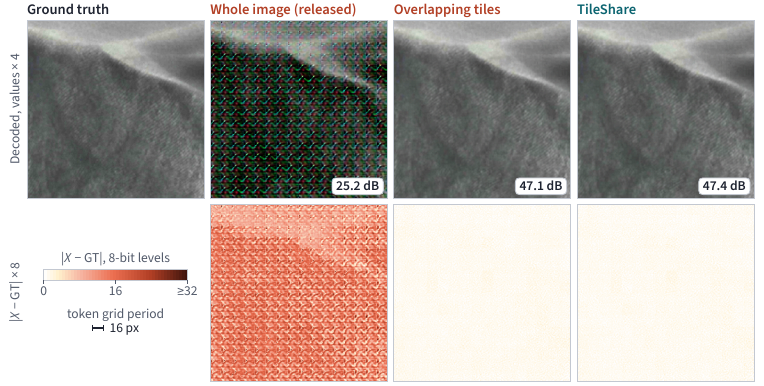}
\caption{\textbf{ViTok-v2 at $4096^2$}: DIV8K image 0880 at the $256\times256$ window where tiled coding gains most over whole-image coding, chosen by the preregistered rule; it is the most extreme case, not a typical one (on average, the gain is 2.0\,dB). Top: decoded crops with local PSNR, displayed with values $\times4$ because the window is dark; bottom: absolute error to the ground truth, computed on the original values ($\times8$). Whole-image coding (released usage) shows a 16\,px token grid; overlapping tiles and \splitk{0} match the ground truth.}
\label{fig:s_vitok}
\end{figure*}

\paragraph{Results in detail.} \Cref{tab:s_vitok} lists every variant and \cref{tab:s_vitok_time} the timings. All $k\le4$ behave alike, and \splitk{8} differs from tiled decoding by $+0.05$\,dB; running \method in the encoder as well adds about 0.04\,dB and shortens the pipeline by a further 5--10\%. Peak memory at $4096^2$ is 2.87\,GB for \splitk{0}, 4.05\,GB for tiled and 3.12\,GB for whole-image decoding (5B: 14.1, 17.9 and 14.8\,GB); measured speed-ups reach 91--97\% of the FLOP bounds. Whole-image coding shows a 16\,px token grid: the gradient excess on grid columns is $+0.86$ on DIV8K at $4096^2$ against $-0.38$ for tiled coding (\cref{fig:s_vitok}); on one Commons image, a macro photograph of a shell, it fails outright ($-14$ to $-20$\,dB). Hard stitching shows no excess gradient at tile switches on this model. The lower LPIPS and DISTS of \method are consistent across images (37--40 of 40 on Commons-40) and are not accompanied by a loss of high-frequency energy, which is slightly higher than for tiled decoding (0.605 vs.\ 0.594 on Commons-40 at $4096^2$).

\section{Scope}
\label{supp:scope_summary}
Among public ViT decoders, MiniMax-H3 and MAGI-1 ship with overlapping tiles; for ViTok-v2 we built the tiled pipeline (\cref{supp:others}). The speed-up follows the cost bound of Sec.~3.3 of the main paper: it is smaller where attention takes a larger share of the FLOPs (MAGI-1: $1.13\times$) and, end to end, scales with the decoder's share of a request (\cref{supp:eff}). On MAGI-1, keeping a clip's token states resident raises peak memory by 5--9\,GB. Equivalence is tested in PSNR, LPIPS and DISTS; Sec.~5.4 of the main paper reports where the remaining differences concentrate. The per-tile depth is measured once per model and deployment geometry.